%% file: main.tex
\documentclass{article}
\usepackage[utf8]{inputenc}
\usepackage[letterpaper,margin=1in]{geometry}
\usepackage[parfill]{parskip}
\usepackage{xr-hyper,refcount}
\usepackage{graphicx}
\usepackage[utf8]{inputenc} 
\usepackage[T1]{fontenc}    

\usepackage[hyphens, spaces, obeyspaces]{url}     
\usepackage{hyperref}

\hypersetup{
    colorlinks=true,
    linkcolor=blue,
    filecolor=magenta,      
    urlcolor=blue,
}

\usepackage{booktabs}       
\usepackage{amsfonts}       
\usepackage{nicefrac}       
\usepackage{microtype}      
\usepackage[parfill]{parskip}
\usepackage{algorithm,algorithmic}
\usepackage{amsmath,amsthm,amssymb}
\usepackage{dsfont}
\usepackage{mathtools}
\usepackage{cases}
\usepackage{comment}
\usepackage{algorithm,algorithmic}
\usepackage{color}
\usepackage{appendix}
\usepackage{xspace}

\usepackage[english]{babel}
\usepackage{authblk}

\usepackage{libertine}
\usepackage[T1]{fontenc}

\usepackage[round]{natbib}

\usepackage{hyperref}       %
\usepackage{url}            %
\usepackage{booktabs}       %
\usepackage{amsfonts}       %
\usepackage{nicefrac}       %
\usepackage{microtype}      %
\usepackage{xcolor}         %
\usepackage[shortlabels]{enumitem}

\usepackage{amsmath, amssymb, amsfonts}

\newcommand{\smallE}[1]{\mathbb{E}[#1]}
\newcommand{\smallCov}[1]{\text{Cov}(#1)}
\newcommand{\smallVar}[1]{\text{Var}(#1)}
\newcommand{\E}[1]{\mathbb{E}\left[#1\right]}
\newcommand{\diag}[1]{\text{diag}\left(#1\right)}
\newcommand{\Cov}[1]{\text{Cov}\left(#1\right)}

\newcommand{\Xt}{\widetilde{X}}

\usepackage{amsthm}
\usepackage{bbm}
\usepackage{algorithm,algorithmic}
\usepackage{cancel}
\allowdisplaybreaks

\usepackage{graphicx}
\usepackage{tikz}
\usepackage{subcaption}
\usetikzlibrary{shapes,decorations,bayesnet}
\usetikzlibrary{shapes.geometric}
\usetikzlibrary{arrows}
\tikzset{
    vertex/.style={circle,draw,minimum size=1.5em},
    edge/.style={->,> = latex'}
}
\usepackage{multicol}
\usepackage{todonotes}

\newtheorem{theorem}{Theorem}[section]
\newtheorem{lemma}[theorem]{Lemma}
\newtheorem{proposition}{Proposition}
\newtheorem{remark}{Remark}

\newtheorem{namedexample}{Example}
\newtheorem{definition}{Definition}

\title{
Domain Adaptation under Missingness Shift
}
\date{}

\author{Helen Zhou, Sivaraman Balakrishnan, Zachary C. Lipton\\
Carnegie Mellon University\\
\text{\{
\href{mailto:hlzhou@andrew.cmu.edu}{hlzhou},
\href{mailto:sbalakri@andrew.cmu.edu}{sbalakri},
\href{mailto:zlipton@cmu.edu}{zlipton}\}@andrew.cmu.edu}}

\begin{document}

\maketitle
\begin{abstract}
\input{sections/00_abstract}

\end{abstract}
\section{Introduction}
\input{sections/10_intro}

\section{Related Work}
\label{sec:related_work}
\input{sections/20_related_work}

\section{DAMS Problem Setup}
\label{sec:missshift_definition}
\input{sections/30_setup}

\section{Cost of Non-Adaptivity}
\label{sec:nonadaptivity}
Here, we provide intuition on the cost of not adapting the source predictor to the target domain in DAMS with UCAR.
\input{sections/35_nonadaptivity_cost}

\input{sections/36_indicators}

\input{sections/37_l2_reg}

\section{Identification Results}
\label{sec:identification_wo_indicators}
\input{sections/40_identification}
\clearpage
\section{Estimation Results}
\label{sec:estimation_wo_indicators}
\input{sections/50_estimation}

\section{Experiments}
\label{sec:experiments}
\input{sections/60_experiments}

\section{Discussion}
\label{sec:conclusion}
\input{sections/70_conclusion}

\bibliographystyle{apalike}
\bibliography{refs2}

\newpage
\onecolumn
\appendix
\input{sections/80_appendix}

\end{document}

%% file: sections/00_abstract.tex
Rates of missing data often depend on 
record-keeping policies and thus 
may
change across times and locations,
even when the underlying features are comparatively stable. 
In this paper, we
introduce
the problem of
Domain Adaptation under Missingness Shift (DAMS).
Here, (labeled) source data 
and (unlabeled) target data would be exchangeable 
but for different 
missing data mechanisms.
We show that if missing data indicators 
are available, 
DAMS reduces to 
covariate shift.
Addressing cases where such indicators are absent, 
we establish the following theoretical results
for underreporting completely at random:
(i) covariate shift is violated
(adaptation is required);
(ii) 
the optimal linear source predictor
can perform arbitrarily
worse on the target domain
than always predicting the mean;
(iii) the optimal target predictor can be identified,
even when the missingness rates themselves are not;
and (iv) for linear models,
a simple analytic adjustment 
yields consistent estimates
of the optimal target parameters. 
In experiments on synthetic and semi-synthetic data,
we demonstrate the promise of our methods when assumptions hold.
Finally, we discuss a rich family of future extensions.

%% file: sections/10_intro.tex
As of October 2021, following extensive awareness campaigns 
and mass distribution efforts
promoting COVID-19 vaccines, 
approximately 79.2\% of the U.S. population 
over age 18 
had received at least one dose \citep{cdc_vaccination_rates}.
And yet, when collaborating with a 
regional healthcare provider, 
we found only 40.5\% of 
121,329 
adults
tested for COVID-19 
were tagged
indicating 
positive vaccination status
in the electronic medical record (EMR).
This was not a regional anomaly---cross referencing 
with vaccination data from the CDC, 
between 75.7\% and 90.3\% 
of the 
adult population in the region 
had 
actually 
received at least one dose. 
A more plausible explanation is that many patients 
were
vaccinated outside of the hospital system
(e.g., at a pharmacy or football stadium) 
but that this information was never reported
to the hospital system and thus never captured 
in the EMR.

Now suppose 
that 
our collaborator 
decided to 
update 
their intake form
to include a question about 
vaccination status. 
Overnight, the 
rate
of 
patients 
being 
tagged in the EMR as vaccinated would increase dramatically.
Absent any shift in the actual health status of patients,
the distribution of observed data
would still shift, 
owing 
to this 
sudden 
change in clerical practices. 
In real-world 
healthcare 
settings,
such changes in missingness rates are common.
Furthermore, 
as in our 
vaccination 
example,
indicators disambiguating which features
are genuinely negative (vs. missing)
cannot be taken for granted.
Faced with data from different time periods or locations,
each characterized by different patterns of missing data,
how should machine learning (ML) practitioners 
leverage the available data 
to 
get 
the best possible predictor
on a target domain?
While missing data and formal models of distribution shift
are both salient concerns of the ML community,
no work to date provides guidance 
on how to adjust a predictor under such shocks.

In this work, we introduce \emph{missingness shift}, 
where distributional shocks arise due
to changes in the pattern of missingness 
(Figure \ref{fig:dgp}). 
In this setup, all domains share a fixed 
underlying distribution $P(X,Y)$,
and observed covariates $\widetilde{X}$
are produced by stochastically zeroing out 
a subset of the underlying \emph{clean} covariates,
i.e., each $\widetilde{X} = X \odot \xi$ 
for some $\xi \in \{0,1\}^d$.
We propose the \textbf{Domain Adaptation under Missingness Shift} problem,
where the learner aspires 
to recover the optimal target predictor 
given \emph{labeled} data from the source distribution $P^s(\widetilde{X},Y)$,
and unlabeled data from the target (deployment) distribution $P^t(\widetilde{X})$.

We focus primarily on a special DAMS setting 
where the components of $\xi$'s (one per feature) 
are drawn from independent Bernoullis 
with unknown constant probabilities. 
First, we show that when missingness indicators $(1-\xi)$ are available, 
missingness shift is an instance of covariate shift. 
However, absent indicators,
missingness shift constitutes 
neither covariate shift nor label shift.
Thus, adaptation is required.
We demonstrate that under DAMS, 
the optimal source predictor 
may even 
exhibit arbitrarily higher MSE
than just guessing the label mean $\mathbb{E}[Y]$.
One natural strategy might be to 
relate 
the source and target distributions 
to the underlying clean distribution, 
which we show is identified
when missingness rates are known.
However, we show that
the missingness rates are not, in general, identifiable.
Fortunately, as we prove,
the target distribution 
(and thus optimal target predictor)
is nevertheless identifiable, 
requiring only that we estimate the (observable) relative
proportions of nonzero values for each covariate across domains.
Using these relative proportions,
we derive a simple adjustment formula that yields
the optimal linear predictor on the target domain.
Additionally, 
we provide a non-parametric, model-agnostic procedure which attempts to transform source data into labeled data i.i.d. to the target distribution.
Finally, we confirm the validity of our approach
and demonstrate empirical gains
in settings where our assumptions hold
through synthetic and semi-synthetic experiments.

%% file: sections/20_related_work.tex
There is a rich history of learning 
under various missing data mechanisms when missing data indicators are available
\citep{rubin1976inference,robins1994estimation,little2019statistical,gelman2020regression}. 
Common practices for handling missing data
include discarding all samples with missingness
(complete-case analysis) \citep{little2019statistical}, 
imputing with mean or last value carried forward, 
combining inferences from multiple imputations \citep{rubin1996MI,van2011mice}, 
matching-based algorithms, 
iterative regression imputation \citep{stekhoven2012missforest,NEURIPS2021_5fe8fdc7}, building missingness indicators into model architecture \citep{le2020neumiss},
and including missingness indicators as features \citep{groenwold2012missing,lipton2016modeling, little2019statistical}. 
However, these techniques require indicators for whether each covariate is missing in the first place.

In single cell RNA sequencing, missing data indicators 
are often absent
in count data due to dropout, where 
a tiny proportion
of the transcripts
in each cell are sequenced, 
so
expressed transcripts can go undetected
and are instead assigned a zero value. 
This is often dealt with by leveraging domain-specific knowledge to inform probabilistic models, such as 
assuming a zero-inflated negative binomial distribution of counts~\citep{risso2018general},
using a mixture model to identify likely missing values
before imputing 
with nonnegative least squares regression~\citep{li2018accurateSCIMPUTE}, 
adopting a Bayesian approach to estimate 
a posterior distribution of gene expressions~\citep{huang2018saver},
or 
graph-based methods
on a lower dimensional manifold
derived from principal component analysis~\citep{van2018recoveringMAGIC}.

In survey data, underreporting (i.e. missingness without indicators) arises in binary data when respondents give false negative responses to questions.
As noted in \citet{SECHIDIS2017159}, this 
can be viewed as a form of misclassification bias. 
In its simplest form, an underreported variable 
has \emph{specificity} $p(\widetilde{x} = 0 | x = 0) = 1$
and \emph{sensitivity} $p(\widetilde{x} = 1 | x = 1) < 1$ 
(one minus the rate of missingness).
If %
sensitivity is independent of 
outcome $Y$,
this is referred to as \emph{non-differential} misclassification, which often, but not always
biases measures of association towards zero
\citep{dosemeci1990does,brenner1994varied}.
Given 
knowledge of 
the specificities and sensitivities, 
prior work has derived adjusted estimators 
for the log-odds ratio \citep{chu2006sensitivity}
and relative-risk \citep{rahardja_young_2021}
under non-differential exposure misclassification.
Recent work has also provided conditions 
under which the joint distribution $p(y, \widetilde{a} | x)$ 
(outcome $y$, single binary underreported exposure $\widetilde{a}$, 
and fully observed covariates $x$) is identifiable \citep{adams2019learning}. 

In our setting, for binary covariates,
estimating the missingness rates takes the form 
of learning from positive and unlabeled data
\citep{elkan2008learning, bekker2020learning}.
Here, identification of the missingness rates
hinges on the existence of a separable positive subdomain~\citep{garg2021mixture},
which may not hold in problems of interest. 
Many canonical distribution shift problems address
adaptation under different forms of structure, 
including covariate shift
\citep{shimodaira2000improving,zadrozny2004learning,huang2006correcting,sugiyama2007direct,gretton2009covariate}, label shift %
\citep{saerens2002adjusting, storkey2009training,zhang2013domain,lipton2018detecting,garg2020unified},
and concept drift \citep{tsymbal2004problem,gama2014survey}.
We show that missingness shift with missing data indicators can often be reinterpreted as a form of covariate shift, but to our knowledge, missingness shift without indicators 
does not fit neatly into any previous setting.

%% file: sections/30_setup.tex
\begin{figure}
    \centering
    \includegraphics[width=0.45\textwidth]{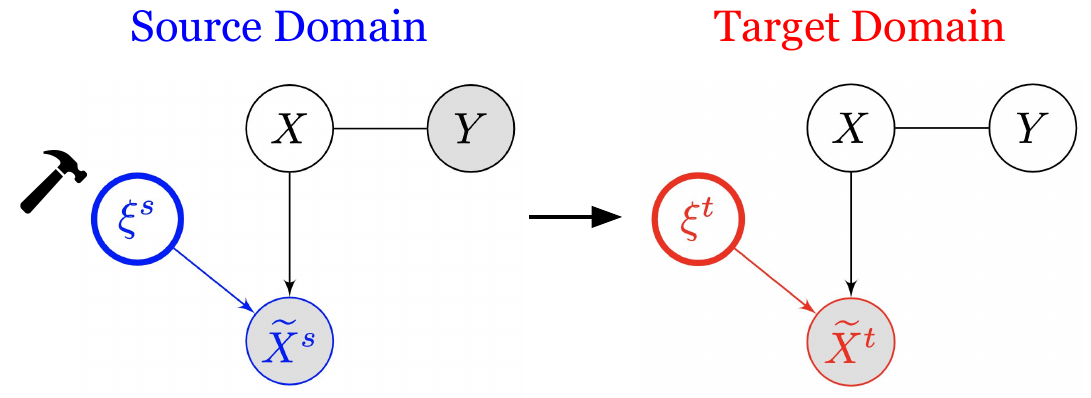}
    \caption{DAMS with UCAR.
    The source and target data are drawn from the same $P(X,Y)$, but differ in how $\xi$ (and hence $\widetilde{X}$) takes its value. Shaded nodes are observed.
    Observed covariates are generated as $\widetilde{X} = X \odot \xi$. 
    The undirected edge between X and Y indicates that they can have an arbitrary bidirectional relationship.
    }
    \label{fig:dgp}
\end{figure}

First, we define 
notation for 
(1) missing data;
(2) missingness shift; and 
(3) the DAMS problem.
Then, motivated by the medical setting, 
we focus on a specific form of DAMS
(Figure \ref{fig:dgp}) for the remainder of the paper.

Let us denote \emph{clean} covariates $X \in \mathbb{R}^d$ and labels $Y \in \mathbb{R}$. Let $X_j$ denote the $j$th covariate, for $j \in \{1, 2, ..., d\}$.

\paragraph{Missing Data}
In every environment $e$ with missing data, we 
do not directly observe $X$, but instead 
observe \emph{corrupted} covariates: 
$$\widetilde{X} = X \odot \xi,$$
where $\xi \in \{0, 1\}^d$ and $(X, Y, \xi) \sim P^e$ for distribution $P^e$.
Note that mask $\xi$ is the complement of missing data indicators $(1-\xi)$.
In this paper, we assume no missingness in $Y$ in labeled data.
An important assumption of missing data problems is how $\xi$ takes its value, e.g. independent of other covariates, dependent on other covariates, etc. Furthermore, $\xi$ may or may not be observed.

\begin{definition}[Missingness Shift]
Consider a source domain $s$ and target domain $t$ in which $X$ and $Y$ are drawn from the same underlying distribution, i.e. $P(X,Y) = P^s(X,Y) = P^t(X,Y)$. 
\emph{Missingness shift} occurs when the missing data mechanism differs between $s$ and $t$, i.e. $P^s(\xi | \cdot) \neq P^t(\xi|\cdot)$.
\end{definition}

\paragraph{Domain Adaptation under Missingness Shift}
Suppose missingness shift occurs between source domain $s$ and target domain $t$.
Given 
observations of 
corrupted labeled source data 
$\{(\widetilde{X}^{s,i}, Y^{s,i})\}_{i=1}^{n_s}$ where 
$(\widetilde{X}^{s, i}, Y^{s, i}) \sim P^s(\widetilde{X}, Y)$,
as well as
corrupted unlabeled target data 
$\{\widetilde{X}^{t, i}\}_{i=1}^{n_t}$ where 
$\widetilde{X}^{t, i} \sim P^t(\widetilde{X})$,
the goal of DAMS is to learn an optimal predictor
on the corrupted target domain data.
In this paper, we focus on regression-type tasks, where optimality is measured by the squared error on the corrupted target domain data, and we seek the optimal predictor $\mathbb{E}_{(\widetilde{X}^t,Y) \sim P^t}[Y | \widetilde{X}^t] $.

As we will show (in Section \ref{sec:nonadaptivity}), DAMS is particularly challenging when missing data indicators are \emph{not available}.
This setting without observing $\xi$ is trickiest when there are 
a substantial number of 
true 0 values that now become indistinguishable from missing values.
Without knowledge of which data are missing versus true 0s, conventional techniques for imputing missing entries do not apply. 
To make this difficult setting tractable, we define the DAMS with underreporting completely at random (UCAR) setting, which we focus on in this paper.

\paragraph{DAMS with UCAR}
Assume that $\xi$ (unobserved) is drawn independently of other variables, and parameterized by \emph{constant (but unknown) missingness rates}
$m^s \in [0,1]^d$ in source and $m^t \in [0,1]^d$ in target. That is, $\forall j \in \{1, 2, ..., d\}$, we have independently drawn
$\xi^s_{j} \sim \text{Bernoulli}(1 - m^s_{j})$ and 
$\xi^t_{j} \sim \text{Bernoulli}(1 - m^t_{j})$, abbreviated as:
\begin{align*}
    \xi^s &\sim \text{Bernoulli}(1 - m^s)\\
    \xi^t &\sim \text{Bernoulli}(1 - m^t).
\end{align*}
For binary data, this setting without missingness indicators is known as \emph{underreporting}. We thus refer to this setting as underreporting completely at random, but note our results are not limited to binary data.

%% file: sections/35_nonadaptivity_cost.tex
Let us start with a
simple motivating example. 
Define the 
risk of an estimator $\widehat{h}$ to be $r(\widehat{h}) = \smallE{(Y - \widehat{h}(X))^2}$.

\begin{namedexample}[Redundant Features]\label{example:redundant_features}
Let $m^s = [1-\epsilon, \epsilon]$ and $m^t = [\epsilon, 1-\epsilon]$. Consider the data generating process:
\begin{equation*}
\begin{aligned}[c]
Z &= u_Z\\
X_1 &= Z \\
X_2 &= Z \\
Y &= Z + u_Y
\end{aligned}
\qquad \qquad
\begin{aligned}[c]
u_Z &\sim \mathcal{N}(0, \sigma^2_z)\\
u_Y &\sim \mathcal{N}(0, \sigma^2_y)
\end{aligned}
\end{equation*}
where $\sigma_z$ is a positive constant, $Z$ is a latent variable, $X_1$ and $X_2$ are observed, and $Y$ is the outcome of interest. 
\end{namedexample}
\begin{remark}
In Example \ref{example:redundant_features}, as $\epsilon \rightarrow 0$, the optimal linear source and target predictors have coefficients $\beta^s_* \rightarrow [0, 1]$ and $\beta^t_* \rightarrow [1, 0]$. 
The risk on target data $r^t(\beta^s_*)\rightarrow \text{Var}(Y)$. 
\end{remark}
That is, failing to adapt to the target 
levels of
missingness results in performance no better than simply guessing the label mean (proof in Appendix \ref{app:motivating_examples}). Now, let us consider a slightly more complex example.

\begin{namedexample}[Confounded Features]\label{example:confounded}
Now, suppose that $m^s = [0, 0]$ and $m^t = [1, 0]$. For some constants $a, b, c$ consider the following data generating process:
\begin{equation*}
\begin{aligned}[c]
X_1 &= \nu_{1}\\
X_2 &= aX_1 + \nu_{2}\\
Y &= bX_1 + cX_2 + \nu_Y
\end{aligned}
\qquad \qquad
\begin{aligned}[c]
\nu_{1} &\sim \mathcal{N}(0, 1)\\
\nu_{2} &\sim \mathcal{N}(0, 1)\\
\nu_Y &\sim \mathcal{N}(0, 1)
\end{aligned}
\end{equation*}
\end{namedexample}
\begin{remark}
In Example \ref{example:confounded}, 
the optimal source and target predictors are $\beta^s_* = [b, c]$ and $\beta^t_* = [0, \frac{ab}{a^2 + 1} + c]$.
By setting $a = -\frac{b}{c}$, we can show that
for any $\tau > 1$, 
there exists values of $a, b, c$ such that $r^t(\beta_*^s) > \tau \text{Var}(Y)$.
\end{remark}

Here, failing to adapt to target levels of missingness can result in performance arbitrarily \emph{worse} than predicting the constant label mean (proof in Appendix \ref{app:motivating_examples}).

%% file: sections/36_indicators.tex
\paragraph{Observing $\xi$, Reduction to Covariate Shift}\hspace{0.5em}~In DAMS with UCAR, missing data indicators are absent. By contrast, \emph{suppose we observed missingness indicators} $(1 - \xi)$ (and hence $\xi$). 
Then, we show that missingness shift is an instance of  
covariate shift, where the optimal predictor does not change across domains. 
This 
result 
holds not only when $\xi$ is drawn independently of other covariates, but also when it is dependent on 
other 
completely observed covariates (proof in Appendix \ref{app:covariate_shift}).
Here, when $\xi$ is ``drawn independently of other covariates,'' 
as described in the DAMS with UCAR setup (Section \ref{sec:missshift_definition}), 
we have that $\xi \sim \text{Bernoulli}(1 - m)$ for some constant vector of missingness rates $m \in [0, 1]^d$.
When $\xi$ is drawn depending only on other completely observed covariates, we have that 
some subset of covariates $X_c \subseteq X$ is completely observed (i.e. no missingness), and the missingness of the other covariates $X_m = X \setminus X_c$ depends on $X_c$. That is, $\xi \sim \text{Bernoulli}(f(X_c))$ for some function $f: \mathbb{R}^{|X_c|} \rightarrow [0, 1]^{|X_m|}$.  \citet{mohangraphical} classifies these missingness mechanisms as MCAR (missing completely at random) and v-MAR (variant of the missingness at random described by \citet{rubin1976inference}), respectively.

\begin{proposition}[Reduction to Covariate Shift]\label{prop:covshift}
Assume we observe $\xi$. Consider augmented covariates $\tilde{x}' = (\tilde{x}, \xi)$. 
When $\xi$ is drawn independently of other covariates or depending only on other completely observed covariates, missingness shift satisfies the covariate shift assumption,
i.e, $P^s(Y|\widetilde{X}' = \tilde{x}') = P^t(Y|\widetilde{X}' = \tilde{x}')$.
\end{proposition}

Covariate shift problems are well-studied \citep{shimodaira2000improving,zadrozny2004learning,huang2006correcting,sugiyama2007direct,gretton2009covariate}.
When source and target distributions have shared support, 
covariate shift only requires adaptation
under model misspecification \citep{shimodaira2000improving},
where the most common approach is to re-weight examples
according to $p^t(x)/p^s(x)$, rendering
the (re-weighted) training and target data exchangeable. 
However, even given missingness indicators,
DAMS may still require some care. 
For example, in the augmented covariate space (with missing data indicators), one might need more complex models than in the original covariate space.
When re-weighting is necessary, 
the structure of the DAMS problem 
might be leveraged to estimate importance weights more efficiently,
or to identify the optimal target predictor
in certain cases where missingness introduces non-overlapping support. 
However, because our work is primarily motivated
by underreporting in the medical setting,
we focus our attention on the case 
where missingness indicators are absent.

%% file: sections/37_l2_reg.tex
\paragraph{UCAR as Regularization}\hspace{0.5em}~While the optimal predictor does not change across domains when $\xi$ are observed (as the covariate shift assumption holds),
it is less obvious \emph{how missingness without indicators impacts the optimal predictor}.
To build intuition on the effect of 
underreporting completely at random, we note that applying mask $\xi$, which zeros out covariates with some probability, resembles the mechanism of dropout in neural networks. 
Using similar theoretical arguments as in how dropout acts as a form of regularization \citep{NIPS2013_38db3aed_dropout}, we show that for linear models, the phenomenon of UCAR in data with constant missingness rate $m$ translates into a form of regularization on the resulting predictor (proof in Appendix \ref{sec:l2_reg_approx}). First, we show that for generalized linear models, UCAR results in a regularization effect. Here, generalized linear models are defined as $p_\beta(y|x) = h(y)\exp\{yx\cdot \beta - A(x \cdot \beta)\}$, where $h(y)$ is a quantity independent of $x$ and $\beta$, and $A(\cdot)$ is the log partition function, and the negative log likelihood objective is $l_{x,y}(\beta) = -\log p_\beta (y|x)$. Then, considering linear regression, we show that the regularization penalty can be viewed as a form of L2 regularization.

\begin{theorem}
\label{theorem:l2_reg_approx}
Under UCAR with missingness rates $m \in [0, 1)^d$, 
the minimizer $\widehat{\beta}$ of
the negative log likelihood of the corrupted training data 
$\widetilde{X}$
scaled by $\frac{1}{1-m}$ is given by:
\begin{align*}
    \widehat{\beta} 
    &= \arg\min_{\beta \in \mathbb{R}^d} \sum_{i=1}^n \mathbb{E}_{\xi}[l_{\widetilde{x}^{(i)}, y^{(i)}}(\beta)] \\
    &= \arg\min_{\beta \in \mathbb{R}^d} \sum_{i=1}^n l_{x^{(i)}, y^{(i)}}(\beta) + R(\beta),
\end{align*}
where $l_{\widetilde{x}^{(i)}, y^{(i)}}(\beta)$ and $l_{x^{(i)}, y^{(i)}}(\beta)$ are the negative log likelihoods of a corrupted sample and the corresponding clean sample (respectively). For linear regression, the regularization term $R(\beta)$ is given by:
$$R(\beta) = \frac{1}{2} \left (\beta \widetilde{\Delta}_{\text{diag}} \right )^\top \left (\beta \widetilde{\Delta}_{\text{diag}} \right ),$$
where we define $\widetilde{\Delta}_{\text{diag}} = \text{diag}\left ( \sqrt{\frac{m}{1 - m}} \right ) \text{diag}(I)^{1/2}$, 
where $\text{diag}\left(\sqrt{\frac{m}{1-m}}\right)$ refers to a diagonal matrix with $\sqrt{\frac{m_j}{1-m_j}}$ on the diagonal, and $\text{diag}(I)^{1/2}$ refers to the square root of the diagonal of the Fisher information matrix.
\end{theorem}

Thus, for linear regression, applying missingness rates $m$ to data scaled by $\frac{1}{1-m}$ can be viewed as a form of L2 regularization of $\beta$ scaled by  $\widetilde{\Delta}_{\text{diag}}$.

%% file: sections/40_identification.tex
This section shows that in DAMS with UCAR, the clean joint distribution $p$ is identifiable from the corrupted joint distribution $\widetilde{p}$ with missingness rates $m \in [0,1)^d$ when $m$ is known (Lemma \ref{lemma:clean_from_corrupted}). However, 
$m$ is not in general identifiable directly from the observed corrupted data (Remark \ref{remark:unidentifiable_missingness}). 
Instead, we identify \emph{relative} rates of non-missingness from the corrupted data across domains (Remark \ref{remark:relative_missingness}), which can in turn be used to identify the labeled target distribution $\widetilde{p}^t$ from the labeled source distribution $\widetilde{p}^s$ (Theorem \ref{theorem:s_to_t_elementwise}).

First, we define some notation useful for our identification results. Consider vectors $a \in \mathbb{R}^d$ and $b \in \mathbb{R}^d$. Let $a \prec b$ denote that $\forall j \in \{1, 2, ..., d\}$, we have $a_j < b_j$. Similarly, let $a \succeq b$ denote that $\forall j \in \{1, 2, ..., d\}$, $a_j \geq b_j$.

To help clarify the relationship between corrupted and clean distributions, we define the notion of m-reachability. 

\begin{definition}[m-reachable]
We say $b$ is m-reachable from $a$ (denoted $a \leadsto b$) if $\text{ }\exists\text{\hspace{0.1em}}\xi \in \{0,1\}^d$ such that $b = a \odot \xi$.
\end{definition}

\begin{remark}[Characteristics of m-reachability]\label{remark:mreach_characters}
If $a \leadsto b$, then the dimensions of $a$ that are 0 must be a subset of the ones that are 0 in $b$. Additionally, any dimensions that are nonzero in both $a$ and $b$ must match in value.
\end{remark}

For example, if we observe a data point $b = [1, 1, 1]$, the only data point $a$ for which $a \leadsto b$ is $a = [1,1,1]$. If $b = [1,1,0]$, then possible values of $a$ are $a = [1,1,c]$ for any value of $c \in \mathbb{R}$. In binary data, $a \leadsto b \iff a \succeq b$.

Let $p_{x,y} = P(X= x, Y=y)$ denote the probability of some set of covariates $x \in \mathbb{R}^d$ and label $y \in \mathbb{R}$ in the clean distribution, and let $\widetilde{p}_{x,y} = P(\widetilde{X}= x, Y=y)$ denote the same in the corrupted distribution. Throughout the paper we use notation for discrete $X$, but note that it is straightforward to extend the results to continuous $X$ (e.g. by replacing summations with integrals, etc.).
Summing over all possible values of $z \in \mathbb{R}^d$ from which 
$x$ is m-reachable,
$\widetilde{p}$ can be expressed in terms of $p$ and $m$:
\begin{equation}\label{eqn:elementwise_corruption}
    \widetilde{p}_{x,y} = \sum_{z: z \leadsto x} p_{z, y} \cdot \prod_{j=1}^d (1-m_{j})^{[x_j]_{\neq 0}} m_{j}^{[z_j]_{\neq 0} - [x_j]_{\neq 0}}
\end{equation}
where $[ x ]_{\neq 0} \overset{\Delta}{=} \mathds{1}[x \neq 0]$ 
is
an indicator function for nonzero values. 
While it is obvious that one can obtain $\widetilde{p}$ from $p$, we show, surprisingly,
that the above %
system is in fact invertible.

\begin{lemma}\label{lemma:clean_from_corrupted}
Given $m$, where $m \prec 1$, the clean distribution $p$ is identifiable from the corrupted distribution $\widetilde{p}$.
\end{lemma}

Roughly, the proof of Lemma \ref{lemma:clean_from_corrupted} (in Appendix \ref{app:clean_from_corrupted}) rearranges 
equation 
\eqref{eqn:elementwise_corruption} and uses Remark \ref{remark:mreach_characters} to observe that any entry $p_{(x,y)}$ 
can be expressed in terms of $\widetilde{p}$, $m$, and entries of $p$ with fewer zeros. Using proof by induction on the number of zeros
(0 to $d$),
one can show that $p$ is identifiable from $\widetilde{p}$.

Returning to the DAMS problem, \emph{given $m^s$ and $m^t$}, one could in theory identify $p$ from $\widetilde{p}^s$ thru Lemma \ref{lemma:clean_from_corrupted}, and then use equation \eqref{eqn:elementwise_corruption} to derive $\widetilde{p}^t$. 
Unfortunately, however, \emph{missingness rates are not in general identifiable} from the observed corrupted data.

\begin{remark}\label{remark:unidentifiable_missingness}
Missingness rates are not in general identifiable directly from corrupted data. 
To see this, consider the following simple counterexample. 
Consider two distinct possible source distributions
$A \sim \text{Bernoulli}(0.5)$ and 
$B \sim \text{Bernoulli}(0.25)$.
Application of missingness with rates $m_A = 0.5$ to $A$ and $m_B = 0$ to $B$ yields identical corrupted distributions $\widetilde{A} \sim \text{Bernoulli}(0.25)$ and $\widetilde{B} \sim \text{Bernoulli}(0.25)$. Thus, the rates are not identifiable.
\end{remark}

While missingness rates are not in general identified given corrupted data from a single domain, one might hope to nevertheless \emph{relate} the missingness rates between source and target domains. For this, we leverage nonzero values. Whereas observed zeros are a mixture of zeroed-out values and true zeros, 
all observed nonzeros
were %
nonzero in the clean data. Thus, the relative proportions of nonzeros are informative of relative \emph{non-missingness rates} $1-m$. For a covariate $X_j$, where $j \in \{1, ..., d\}$, denote the true proportion of nonzeros in the underlying data as $q_j = P(X_j \neq 0)$. Then, the proportion of observed nonzeros in the corrupted data is $P(\widetilde{X}_j \neq 0) = (1-m_j)q_j$. Vectorized, $P(\widetilde{X} \neq 0)  = (1-m) \odot q$.

\begin{remark}\label{remark:relative_missingness}
The ratio between non-missingness rates $1 - m^t$ and $1-m^s$ is given by:
\begin{align}\label{eqn:rel_missing_mult}
\frac{1-m^{t}}{1-m^{s}} = \frac{(1-m^{t}) \odot q}{(1-m^{s}) \odot q} = \frac{P^t(\widetilde{X} \neq 0)}{P^s(\widetilde{X} \neq 0)} \triangleq 1 - r^{s\rightarrow t},
\end{align}
where the divisions are element-wise. Note that the second-to-last expression is estimable from observed data.
\end{remark}

We refer to $r^{s\rightarrow t} = 1 - \frac{1-m^{t}}{1-m^{s}} = \frac{m^t - m^s}{1 - m^s}$ 
as the \emph{relative missingness rates} between $s$ and $t$.
Interestingly, while identification of the 
\emph{clean} distribution from a corrupted distribution 
(Lemma \ref{lemma:clean_from_corrupted})
may be difficult due to unidentifiability of $m^s$ and $m^t$ in general (Remark \ref{remark:unidentifiable_missingness}), we leverage identifiability of $r^{s \rightarrow t}$ to show that 
\emph{adapting} from one corrupted distribution to another corrupted distribution does not require identification of the clean distribution. 
\begin{theorem}\label{theorem:s_to_t_elementwise}
For source and target distributions $\widetilde{p}^s$ and $\widetilde{p}^t$ with unknown missingness rates 
$m^s$ and $m^t$ (respectively), 
where $m^s \prec 1$,
$\widetilde{p}^t$ is identifiable from $\widetilde{p}^s$ given 
$r^{s\rightarrow t}$:
\begin{equation}\label{eqn:s_to_t_elementwise}
    \widetilde{p}^t_{x,y} = \sum_{z: z \leadsto x} \widetilde{p}^s_{z, y} \cdot \prod_{j=1}^d (1-r_{j}^{s \rightarrow t})^{[x_j]_{\neq 0}} (r_{j}^{s \rightarrow t})^{[z_j]_{\neq 0} - [x_j]_{\neq 0}}.
\end{equation}
\end{theorem}
That is, while the precise missingness rates $m^s$ and $m^t$ 
may be unidentifiable in general from corrupted data, 
one can identify relative missingness rates $r^{s\rightarrow t}$ (Remark \ref{remark:relative_missingness}) and use them to directly identify $\widetilde{p}^t$ from $\widetilde{p}^s$ (proof in Appendix \ref{app:id_joint_from_source}), rather than explicitly using the clean distribution as an intermediate step. 
Note that the form of \eqref{eqn:s_to_t_elementwise} matches that of \eqref{eqn:elementwise_corruption}, with missingness rates set to $m = r^{s\rightarrow t}$.

%% file: sections/50_estimation.tex
We discuss estimation of optimal target predictors from labeled source data 
$\{(\widetilde{X}^{s,i}, Y^{s,i})\}_{i=1}^{n_s}$, drawn from $P^s(\widetilde{X}, Y)$
and
unlabeled target data 
$\{\widetilde{X}^{t, i}\}_{i=1}^{n_t}$, drawn from $P^t(\widetilde{X})$.

\paragraph{Non-parametric adjustment procedure for nonnegative relative missingness {} {}}\label{sec:nonparametric}
The parallels between equations \eqref{eqn:s_to_t_elementwise} and \eqref{eqn:elementwise_corruption} suggest an intuitive non-parametric procedure when $m^s \preceq m^t$, so that $r^{s\rightarrow t} \succeq 0$ (Algorithm \ref{alg:nonparametric}). To obtain data distributed identically to $\widetilde{X}^t$, one can sample masks $\xi^{s\rightarrow t}$ with missingness rates $r^{s\rightarrow t}$ and apply them to 
$\widetilde{X}^s$.
Let us define a \emph{missingness filter} applied to each datapoint $x \in \mathbb{R}^d$ as $\nu_{s\rightarrow t}(x) = x \odot \xi^{s \rightarrow t}$, where $\xi^{s\rightarrow t} \sim \text{Bernoulli}(1-r^{s\rightarrow t})$.
When a missingness filter is applied to a dataset, $\xi^{s\rightarrow t}$ is independently drawn for every data point.
A proof showing that 
labeled data $\{(\nu_{s\rightarrow t}(\widetilde{X}^{s,i}), Y^{s,i})\}_{i=1}^{n_s}$ 
are drawn i.i.d. to
$P^t(\widetilde{X}, Y)$
is 
in Appendix \ref{app:alternate_nonparametric_pf}. For any desired model class, we can now train a predictor on this labeled data.
When $m^s \preceq m^t$, we call this adjustment a \emph{proper} adjustment as it yields a predictor trained on data i.i.d. to labeled target data.

When $m^s \npreceq m^t$, i.e. $r^{s\rightarrow t} \nsucceq 0$, it is less obvious what the proper non-parametric adjustment procedure implied by Theorem \ref{theorem:s_to_t_elementwise} might be. 
As a stopgap measure, we experiment with using a missingness filter of rate $ \max\{r^{s\rightarrow t}, 0\}$ (Algorithm \ref{alg:nonparametric}), but call this an improper adjustment as it does not create data i.i.d to the target distribution.

\begin{algorithm}
  \caption{Non-parametric adjustment procedure \newline(proper adjustment when $m^s \preceq m^t$)}
  \label{alg:nonparametric}
  \begin{algorithmic}[1]
    \STATE Compute $\widehat{q}_j^t = \frac{\text{count}\left(\widetilde{x}_j^t \neq 0 \right)}{n_t}$, $\widehat{q}_j^s = \frac{\text{count}\left(\widetilde{x}_j^s \neq 0 \right)}{n_s}$, and $\widehat{r}^{s\rightarrow t} = 1 - \frac{\widehat{q}^t}{\widehat{q}^s}$.
    \STATE Compute $\widetilde{r}^{s\rightarrow t} = \max\{\widehat{r}^{s\rightarrow t}, 0\}$ (element-wise max).
    Note that if $\widehat{r}^{s\rightarrow t} \succeq 0$, then $\widehat{r}^{s\rightarrow t} = \widetilde{r}^{s\rightarrow t}$.
    \STATE Apply a missingness filter with rate $\widetilde{r}^{s\rightarrow t}$ to source data to get 
    $\{(\widetilde{\nu}_{s\rightarrow t}(\widetilde{X}^{s,i}), Y^{s,i})\}_{i=1}^{n_s}$.
    \STATE Fit a predictor on data 
    $\{(\widetilde{\nu}_{s\rightarrow t}(\widetilde{X}^{s,i}), Y^{s,i})\}_{i=1}^{n_s}$.
  \end{algorithmic}
\end{algorithm}

Step 1 of Algorithm \ref{alg:nonparametric} estimates the relative missingness $r^{s\rightarrow t}$ from data. Using Hoeffding's inequality, we show that with high probability, the estimated $\widehat{r}^{s\rightarrow t}$ is close to $r^{s\rightarrow t}$ (proof in Appendix \ref{sec:nonmissing_prop_samples}).

\begin{theorem}\label{theorem:relative_miss_estimation}
With probability at least $1-\delta$,
\begin{small}
\begin{align*}
    \left| \widehat{r}^{s\rightarrow t} - r^{s\rightarrow t} \right| &\leq \frac{1}{\widehat{q}^s}\left(\sqrt{\frac{\log(4/\delta)}{2n_t}} + (1-r^{s\rightarrow t}) \sqrt{\frac{\log(4/\delta)}{2n_s}}\right).
\end{align*}
\end{small}
\end{theorem}

A proper non-parametric adjustment requires $r^{s \rightarrow t} \succeq 0$. 
Next, we derive a closed-form expression
for the optimal linear target predictor 
for any given relative missingness. %

\paragraph{Closed-Form Solution for Optimal Linear Predictor {} {}}\label{sec:closed_form}
Define the optimal predictor as the one that minimizes mean squared error. 
Given observations of source covariates $\widetilde{X}^s$ and their corresponding labels $Y^s$, as well unlabeled target covariates $\widetilde{X}^t$,
we seek the optimal linear predictor $f_*^t(x^t) = \beta^t_* x^t$ for the target domain. Indeed, $\beta^t_*$ can be expressed in terms of quantities estimable from data (proof in Appendix \ref{app:optimal_linear_from_target}).
\begin{proposition} The optimal linear target predictor is given by:
\begin{align}
    \beta^t_* = \smallE{\Xt^{t\top} \Xt^t}^{-1}\left(r^{s\rightarrow t} \odot \smallE{\widetilde{X}^{s\top} Y^s}\right).\label{eq:opt_linear_target_pred}
\end{align}
\end{proposition}
Thus, \emph{without knowing the levels of missingness}, as long as $m^{s} \prec 1$, the optimal linear predictor for the target domain is nevertheless estimable, using target unlabeled data to derive the covariance $\smallE{\Xt^{t\top} \Xt^t}$. As we show in Appendix \ref{app:any_target_missingness}, it is also possible to compute the entries of $\smallE{\Xt^{t\top} \Xt^t}$ using only source data and relative missingness. 
\begin{proposition}
For $i \neq j$, where $i \in \{1, 2, .., d\}$, $j \in \{1, 2, .., d\}$, we have
\begin{align}
\smallE{\Xt^{t\top} \Xt^t}_{ij} 
&= (1-r^{s\rightarrow t}_i)(1-r^{s\rightarrow t}_j)\smallE{\Xt^{s\top} \Xt^s}_{ij}\label{eq:xtt_offdiagonal}\\
\smallE{\Xt^{t\top} \Xt^t}_{ii} 
    &= (1-r^{s\rightarrow t}_i) \smallE{\Xt^{s\top} \Xt^s}_{ii}.\label{eq:xtt_diagonal}
\end{align}
\end{proposition}
Although $\smallE{\Xt^{t\top} \Xt^t}$ could be estimated from either source or target covariates, in practice with finite samples it might be beneficial to utilize both. 
For example, to adjust for sample size of the source and target datasets, one could take a weighted average of the estimates of $\smallE{\Xt^{t\top} \Xt^t}$, where the weights of the source-derived and target-derived estimates are $\alpha_s = \frac{n_s}{n_s + n_t}$ and $\alpha_t = \frac{n_t}{n_s + n_t}$, respectively.
This 
attempts to adjust for the variance of estimation error due to the
different sample sizes,
however it does not account for 
estimation error in
the relative missingness rate. We leave further exploration of these weightings to future work.
Algorithm \ref{alg:linear} describes the estimation procedure for linear models adjusted for the target domain.

\begin{algorithm}[H]
  \caption{Adjusted linear model learning procedure}
  \label{alg:linear}
  \begin{algorithmic}[1]
    \STATE Compute $\widehat{q}_j^t = \frac{\text{count}\left(\widetilde{x}_j^t \neq 0 \right)}{n_t}$, $\widehat{q}_j^s = \frac{\text{count}\left(\widetilde{x}_j^s \neq 0 \right)}{n_s}$,
    and $\widehat{r}^{s\rightarrow t} = 1 - \frac{\widehat{q}^t}{\widehat{q}^s}$ for all $j \in \{1, 2, .., d\}$.
    \STATE Estimate target-based $\widehat{M}^t = \widehat{\mathbb{E}}[\Xt^{t\top} \Xt^t]$ from unlabeled target samples.
    \STATE Estimate source-based $\widehat{M}^s = \widehat{\mathbb{E}}[\Xt^{t\top} \Xt^t]$ by computing for all $i \neq j$, where $i \in \{1, 2, .., d\}$, $j \in \{1, 2, .., d\}$:
    \begin{align*}
        \widehat{M}^s_{ij} 
        &= (1-\widehat{r}^{s\rightarrow t}_i)(1-\widehat{r}^{s\rightarrow t}_j)\widehat{\mathbb{E}}[\Xt^{s\top} \Xt^s]_{ij}\\
        \widehat{M}^s_{ii} 
        &= (1-\widehat{r}^{s\rightarrow t}_i) \widehat{\mathbb{E}}[\Xt^{s\top} \Xt^s]_{ii} 
    \end{align*}
    \STATE Construct a combined weighted estimate of $\widehat{\mathbb{E}}[\Xt^{t\top} \Xt^t]$: $\widehat{M} = \alpha_s \widehat{M}^s + \alpha_t \widehat{M}^t$
    \STATE Estimate $\widehat{\mathbb{E}}[\widetilde{X}^{s\top} Y^s]$ from source samples, and compute 
    \begin{align*}
    \widehat{\beta}^t = \widehat{M}^{-1}\left(\widehat{r}^{s\rightarrow t} \odot \widehat{\mathbb{E}}[\widetilde{X}^{s\top} Y^s]\right).
    \end{align*}
  \end{algorithmic}
\end{algorithm}

%% file: sections/60_experiments.tex
We apply Algorithms \ref{alg:nonparametric} and \ref{alg:linear} 
to synthetic, semi-synthetic, and real data settings. 
We compare the performance of four variations of predictors:
(1) the oracle predictor (Oracle),
trained with target labeled data
and tested on a held-out target test set;
(2) the source predictor (Source), 
trained on source labeled data without any adjustments; 
(3) the closed-form adjustment (Closed-form Adj.) for linear predictors,
given by Algorithm \ref{alg:linear}; 
and (4) the non-parametric adjustment (Non-param. Adj.), 
given by Algorithm \ref{alg:nonparametric}. 
We also do MissForest imputation of both source and target data, 
treating all zeros as missing values, 
and train a source predictor to evaluate on target (Imputed).

In synthetic and semi-synthetic experiments, 
the data is split 4:1:4:1 to create source training, 
source test, target training, and target test sets. 
Different levels of missingness are applied completely at random 
to source and target datasets. Code is provided at \href{https://github.com/acmi-lab/Missingness-Shift}{https://github.com/acmi-lab/Missingness-Shift}.

\noindent\textbf{Synthetic data experiments}\hspace{0.5em}
We draw 10,000 samples from two simple data-generating processes:

\noindent
\begin{minipage}[t]{.5\columnwidth}
\centering
\emph{Scenario 1: ``Redundant Features"}
\begin{align*}
    u_y &\sim \mathcal{N}(0, 1)\\
    Z &\sim \text{Bernoulli}(0.5)\\
    X_1 &= Z\\
    X_2 &= Z\\
    Y &= Z + u_y
\end{align*}
\end{minipage}%
\begin{minipage}[t]{.5\columnwidth}
\centering
\emph{Scenario 2: ``Confounded Features"}
\begin{align*}
    u_{x_2} &\sim \mathcal{N}(0, 1)\\
    u_{y} &\sim \mathcal{N}(0, 1)\\
    X_1 &\sim \text{Bernoulli}(0.5)\\
    X_2 &= \text{expit}(2X_1 + u_{x_2}) \\
    Y &= X_1 - X_2 + u_y
\end{align*}
\end{minipage}

\begin{figure*}
    \centering
    \begin{subfigure}[b]{0.3\textwidth}
    \includegraphics[width=\textwidth]{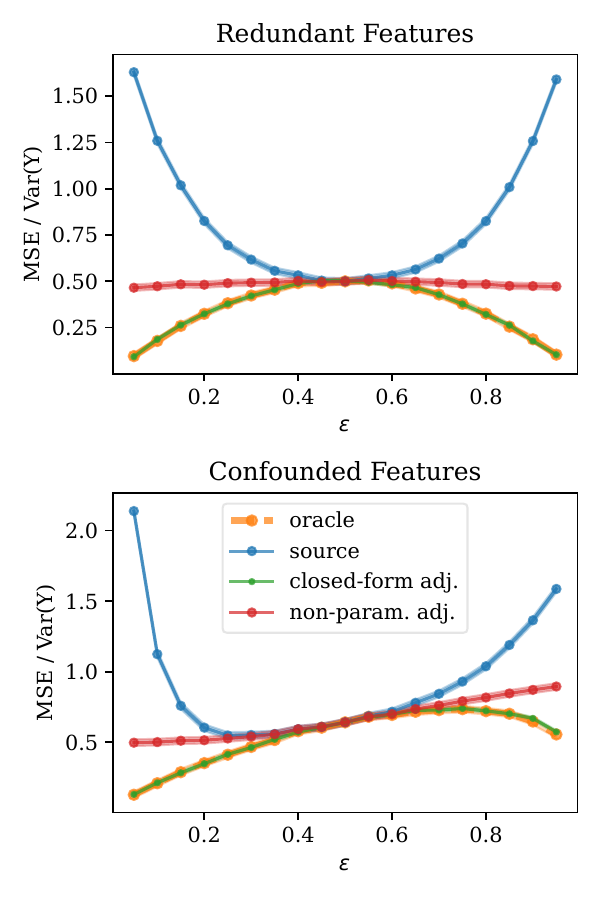}
    \caption{Target domain error of linear models vs. $\epsilon$. Oracle \& closed-form overlap.}
    \label{fig:mclass_perf}
    \end{subfigure}\hfill
    \begin{subfigure}[b]{0.67\textwidth}
    \includegraphics[width=\textwidth]{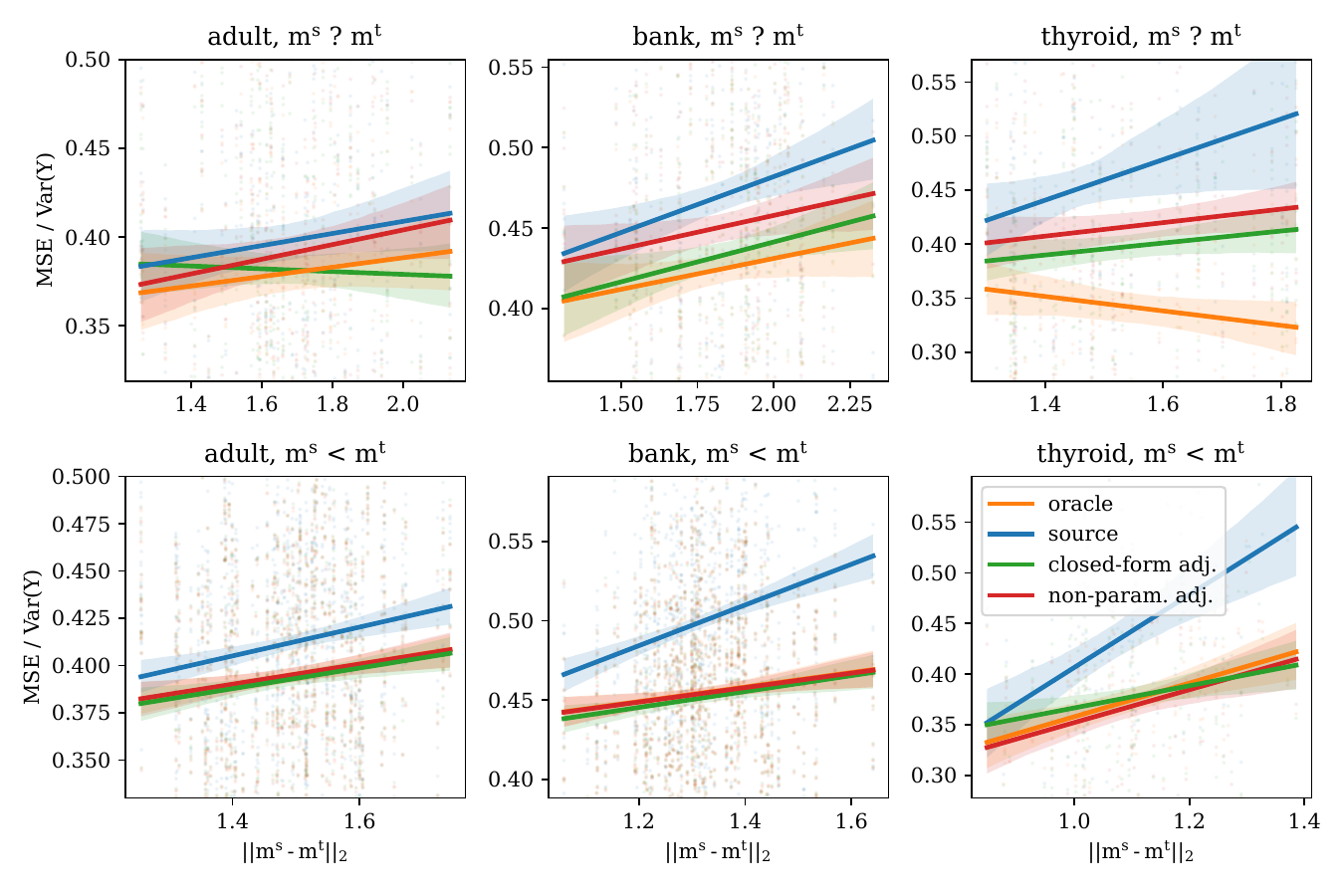}
    \caption{Target domain error of linear models as the L2-norm between $m^s$ and $m^t$ varies. Best-fit line with 95\% confidence intervals from bootstrapping.}
    \label{fig:miss_norm}
    \end{subfigure}
    \caption{$\text{MSE}/\text{Var}(Y)$ of linear models on (a) synthetic and (b) semisynthetic data across varying $m^s$ and $m^t$. }
\end{figure*}

In both, we apply missingness with rates 
$m^s = [1 - \epsilon, \epsilon]$ and $m^t = [\epsilon, 1 - \epsilon]$ 
for varying $\epsilon$ between 0.05 and 0.95 
in increments of 0.05, with 20 runs for each $\epsilon$,
and evaluate the performance of linear predictors (Figure \ref{fig:mclass_perf}). 
At $\epsilon = 0.5$, the source and target domains are identically distributed, 
so Oracle, Source, Closed-form Adj., and Non-param.~Adj.~all 
attain the same mean squared error scaled by variance of the label (MSE/Var(Y)).
As $\epsilon$ approaches 0 or 1, however, the error in the Source predictor 
grows rapidly whereas the Oracle and Closed-form Adj.~errors decrease. 
Since  $m^s \npreceq m^t$, as expected, Non-param.~Adj.~cannot 
fully match the target distribution,
and has intermediate performance.

For $\epsilon = 0.1$, we compare linear regression, XGBoost, and MLP (Table \ref{tab:semisynthetic_perf}). 
In both Scenario 1 and 2, the linear closed-form adjustment 
dramatically outperforms the source linear predictor. 
However, in Scenario 1, source XGBoost and MLP 
almost match the performance of their respective oracles, 
and source XGBoost outperforms the linear oracle. 
On the other hand, in Scenario 2, the linear closed-form adjustment 
outperforms source XGBoost and MLP.

\begin{table*}
\caption{
Target domain $\text{MSE}/\text{Var}(Y)$,
averaged across various missingness levels 
on synthetic and semi-synthetic data.
Confidence intervals are provided in Appendix \ref{app:experiment_details}.
The first two columns are synthetic datasets 
(Redundant Features and Confounded Features),
and the last three columns are semi-synthetic UCI datasets. 
}
\label{tab:semisynthetic_perf}
\setlength{\tabcolsep}{7pt}
\centering

\input{tables/semisynthetic1.tex}
\end{table*}

\begin{table*}[t!]
    \caption{Target domain performance of linear models on eICU 48-hour mortality prediction, where source $s$ and target $t$ can be Hospital 1 (H1) or Hospital 2 (H2). Here, underreporting occurs naturally in the data. Since all features are binary, 
    imputation of all zeros behaves poorly, leading to baseline performance. AUPRC refers to average precision. }
    \centering
    \setlength{\tabcolsep}{8pt}
    \input{tables/eicu2.tex}
    \label{tab:eicu_results}
\end{table*}

\noindent\textbf{Semi-synthetic data experiments}\hspace{0.5em}
Using the adult ($n = 48842$), bank ($n = 48188$), and thyroid binding protein ($n = 2800$) UCI datasets \citep{Dua:2019_UCI}, which contain a mixture of categorical and numerical variables, we construct semi-synthetic datasets by borrowing the covariates, but replacing the labels with synthetically generated labels
that are linear functions of the clean covariates. 
That is, we train using new labels $y_{new} = \beta X$, for randomly sampled $\beta_j \sim \text{Uniform}(0, 10), \forall j\in\{1, 2, ..., d\}$, and original covariates $X$. Source and target missingness rates are sampled under two regimes: 
(1) To test the proper non-parametric adjustment, where $m^s \preceq m^t$,
we sample $m^s_j \sim \text{Uniform}(0, 0.5)$ and $m^t_j \sim m^s_j + (1 - m^s_j)\epsilon$, where $\epsilon \sim \text{Uniform}(0, 0.5)$. 
(2) To simulate a more general form of missingness shift, we sample $m^s_j, m^t_j \sim \text{Uniform}(0, 0.9)$, abbreviated as $m^s \text{ ? } m^t$. For additional experiment and data preprocessing details, see Appendix \ref{app:experiment_details}.

Overall, where adjusted models are applicable/proper,
they perform at least as well as (and often better than) source unadjusted models when compared within each model class (Table \ref{tab:semisynthetic_perf}). 
Among linear models, the closed-form and non-parametric adjustments 
consistently outperform the source predictors.
In nonlinear models, only the non-parametric adjustment applies, 
and this adjustment is only proper if $m^s \preceq m^t$. 
Among nonlinear models, if $m^s \preceq m^t$, either Non-param.~and 
Source tie, or Non-param.~performs best. 
When $m^s \text{ ? } m^t$, Non-param.~(improper adjustment) 
often has the second-best or best performance (especially when no other adjustments apply). 
Ignoring model class, the best-performing model 
for each semi-synthetic dataset is an adjusted model. 
Plotting the line of best fit for MSE/Var(Y) of the linear models 
versus the L2 distance between $m^s$ and $m^t$, 
we note that the Source predictor tends to have 
the stronger positive slope than the Oracle, 
Closed-form Adj., or Non-parametric Adj.~models
(Figure \ref{fig:miss_norm}).

\noindent\textbf{Real data experiments}\hspace{0.5em}
To explore the applicability of our methods to naturally-occurring missingness shifts, we use the FIDDLE 
data 
pre-processing pipeline \citep{tang2020fiddle} on the eICU Collaborative Research Database \citep{pollard2018eicu}, which contains data from critical care units across several hospitals. 
FIDDLE extracts binary feature vectors capturing several patient characteristics, including demographics, physiological measurements, labs, medications, etc. We extract the binary 48-hour mortality outcome for patients in two of the hospitals with the most data ($n_1 = 3006$, $n_2 = 2663$), and verify that the prevalences of the covariates are different across these two hospitals. Additional data and experiment details are provided in Appendix \ref{app:experiment_details}.

We train linear models to predict 
mortality,
and evaluate 
MSE, AUROC, and AUPRC. Since the preprocessed data only contains binary features, MissForest imputation of all zeros results in a dataset consisting entirely of ones, and the linear model learns to simply predict the label mean and only achieves baseline performance.
Estimated 
relative missingness
indicates that $m^s \npreceq m^t$ (Appendix \ref{app:experiment_details}), so the non-parametric estimation procedure is not expected to produce labeled data i.i.d. to the target distribution. The source predictor achieves highest AUROC and AUPRC.

Note, however, that beyond missingness levels, there are also several other aspects of the data distribution that likely differ between these two hospitals. Different hospitals likely have different underlying $P(X, Y)$, and in practice, missingness could be dependent on other covariates (e.g. a doctor may choose not to perform a test based on patient state). Thus, fundamental assumptions of our adaptation methods are likely violated in this dataset.

%% file: tables/semisynthetic1.tex
\begin{tabular}{lcccccccc}
\toprule
{} & \textbf{Rednd.} &  \textbf{Confnd.} &   \multicolumn{2}{c}{\textbf{Adult}}&    \multicolumn{2}{c}{\textbf{Bank}} & \multicolumn{2}{c}{\textbf{Thyroid}} \\
\cmidrule(lr){2-2}\cmidrule(lr){3-3}\cmidrule(lr){4-5}\cmidrule(lr){6-7}\cmidrule(lr){8-9}
{} &   \small{$m^s \text{ ? } m^t$}  & \small{$m^s \text{ ? } m^t$}  & \small{$m^s \preceq m^t$} &   \small{$m^s \text{ ? } m^t$} &    \small{$m^s \preceq m^t$} &    \small{$m^s \text{ ? } m^t$} & \small{$m^s \preceq m^t$} & \small{$m^s \text{ ? } m^t$} \\
\midrule
\multicolumn{9}{c}{\textbf{Linear Regression Models}}\\
\midrule
Oracle           &    \textcolor{gray}{0.178} &     \textcolor{gray}{0.206} &  \textcolor{gray}{0.420} &  \textcolor{gray}{0.362} &  \textcolor{gray}{0.338} &  \textcolor{gray}{0.433} &  \textcolor{gray}{0.298} &  \textcolor{gray}{0.251} \\
Source           &                      1.259 &                       1.103 &                             0.437 &                    0.380 &                             0.371 &                    0.480 &                    0.350 &                    0.320 \\
Imputed          &                      1.002 &                       0.918 &                             0.490 &                    0.483 &                             0.501 &                    0.592 &                    0.306 &                    0.358 \\
Closed-form &             \textbf{0.186} &              \textbf{0.209} &                             0.422 &           \textbf{0.363} &                             0.339 &           \textbf{0.442} &                    0.316 &           \textbf{0.291} \\
Non-param.  &                      0.473 &                       0.492 &                    \textbf{0.420} &                    0.373 &                    \textbf{0.338} &                    0.459 &           \textbf{0.293} &           \textbf{0.291} \\
\midrule
\multicolumn{9}{c}{\textbf{XGBoost Models}}\\
\midrule
Oracle           &  \textcolor{gray}{0.166} &     \textcolor{gray}{0.200} &  \textcolor{gray}{0.398} &  \textcolor{gray}{0.354} &  \textcolor{gray}{0.287} &  \textcolor{gray}{0.453} &  \textcolor{gray}{0.316} &  \textcolor{gray}{0.274} \\
Source           &                    \textbf{0.166} &                       0.475 &           \textbf{0.399} &           \textbf{0.379} &                             0.305 &           \textbf{0.500} &           \textbf{0.310} &           \textbf{0.352} \\
Imputed          &                             1.002 &                       1.157 &                    0.512 &                    0.521 &                             0.492 &                    0.708 &                    0.355 &                    0.441 \\
Non-param.  &                             0.425 &              \textbf{0.473} &           \textbf{0.399} &                    0.392 &                    \textbf{0.287} &                    0.503 &           \textbf{0.310} &                    0.381 \\
\midrule
\multicolumn{9}{c}{\textbf{MLP Models}}\\
\midrule
Oracle           &    \textcolor{gray}{0.166} &     \textcolor{gray}{0.201} &  \textcolor{gray}{0.389} &  \textcolor{gray}{0.343} &  \textcolor{gray}{0.295} &  \textcolor{gray}{0.458} &  \textcolor{gray}{0.279} &  \textcolor{gray}{0.230} \\
Source           &             \textbf{0.184} &              \textbf{0.321} &                             0.399 &                    0.357 &                    0.322 &                    0.499 &                    0.320 &                    0.303 \\
Imputed          &                      1.003 &                       0.924 &                             0.480 &                    0.468 &                    0.484 &                    0.668 &                    0.304 &                    0.345 \\
Non-param.  &                      0.436 &                       0.470 &                    \textbf{0.389} &           \textbf{0.355} &           \textbf{0.294} &           \textbf{0.487} &           \textbf{0.278} &           \textbf{0.272} \\
\bottomrule
\end{tabular}%

%% file: tables/eicu2.tex
\begin{tabular}{lccccc}
\toprule
         Model Class & $s$ & $t$ &                  MSE &                  AUROC & AUPRC\\
\midrule
\textcolor{gray}{Oracle} &  \textcolor{gray}{H1} &  \textcolor{gray}{H1} & 
\textcolor{gray}{0.103 (0.088 -- 0.117)} &  \textcolor{gray}{0.713 (0.652 -- 0.775)} &  \textcolor{gray}{0.236 (0.156 -- 0.317)} \\
Source &  H2 &  H1 &  0.143 (0.135 -- 0.151) &  \textbf{0.593 (0.563 -- 0.623)} &  \textbf{0.146 (0.122 -- 0.170)} \\
Imputed & H2 & H1 & \textbf{0.089 (0.081 -- 0.097)} &  0.500 (0.500 -- 0.500) &  0.097 (0.088 -- 0.106))\\
Closed-form Adj. &  H2 &  H1 &  0.439 (0.223 -- 0.655) &  0.540 (0.509 -- 0.571) &  0.123 (0.103 -- 0.143) \\
Non-param.~Adj. &  H2 &  H1 &  0.142 (0.133 -- 0.150) &  0.555 (0.537 -- 0.573) &  0.126 (0.108 -- 0.144) \\
\midrule
\textcolor{gray}{Oracle} &  \textcolor{gray}{H2} &  \textcolor{gray}{H2} & 
\textcolor{gray}{0.121 (0.100 -- 0.142)} &  \textcolor{gray}{0.601 (0.528 -- 0.675)} &  \textcolor{gray}{0.167 (0.103 -- 0.230)} \\
Source &  H1 &  H2 &  0.122 (0.113 -- 0.131) &  \textbf{0.576 (0.545 -- 0.608)} &  \textbf{0.144 (0.120 -- 0.169)} \\
Imputed & H1 & H2 & \textbf{0.090 (0.082 -- 0.098)} &  0.500 (0.500 -- 0.500) &  0.099 (0.089 -- 0.109)\\
Closed-form Adj. &  H1 &  H2 &  0.373 (0.327 -- 0.420) &  0.556 (0.523 -- 0.588) &  0.122 (0.104 -- 0.141) \\
Non-param.~Adj. &  H1 &  H2 &  0.196 (0.182 -- 0.210) &  0.511 (0.503 -- 0.520) &  0.109 (0.095 -- 0.123) \\ 
\bottomrule
\end{tabular}%

%% file: sections/70_conclusion.tex
This work introduces the domain adaptation under missingness shift (DAMS) problem, and explores DAMS under the underreporting completely at random (UCAR) assumption.
Our synthetic and semi-synthetic experiments demonstrate 
that when assumptions hold, 
the proposed methods (when applicable/proper), 
tend to outperform or perform at least as well 
as unadjusted source predictors in the same model class 
(Table \ref{tab:semisynthetic_perf}).
In linear models, our proposed adjustments
(linear closed-form and non-param. adj.) 
consistently outperform the source predictors,
and sometimes, 
the benefits of adaptation can even outweigh the bias incurred
by restricting to linear models. 
For example, in the Confounded Features, 
Bank $m^s \text{ ? } m^t$, and Thyroid datasets, 
linear adjusted models outperform all Source models, regardless of model class.
Note that even if the underlying relationship 
between clean unobserved covariates $X$ and label $Y$ is linear, 
after $X$ is corrupted by missingness 
to create observed corrupted covariates $\widetilde{X}$,
the new relationship between $\widetilde{X}$ and $Y$ is often nonlinear 
(a phenomenon which has also been noted by \citet{le2020linear}). 
Correspondingly, the best MLP and XGBoost models tend to outperform 
the best linear models (Table \ref{tab:semisynthetic_perf}).

The best-performing model(s) in each of the synthetic and semi-synthetic datasets, 
except for the synthetic Redundant Features dataset, 
use a proposed adjustment (Table \ref{tab:semisynthetic_perf}). 
Although the adjustments perform best in the synthetic Redundant Features dataset 
when restricted to the linear model class, 
the best-performing model in this dataset overall is a source XGBoost model, 
which matches the performance of the oracle. 
In addition to the flexibility of the XGBoost model, 
which improves the oracle XGBoost over the oracle linear model,
a likely reason for improvement of Source XGBoost over Non-param.~Adj.~can 
be found in the particular setup of this scenario. 
Here, $X_1 = X_2 = Z$, and $Y = Z + u_y$, where $u_y \sim \mathcal{N}(0,1)$, 
and so given knowledge of either $X_1$ or $X_2$, 
prediction of $Y$ is straightforward. 
The only applicable adjustment, Non-param.~Adj.~(improper, since $m^s \npreceq m^t$), 
would zero out much of the data to bring the missingness rate in $X_1$ from 0.9 to 0.1, 
thus making prediction harder. 
There are also multiple settings in which Source XGBoost 
performs similarly to Non-param.~Adj.~XGBoost 
(Confounded Features, Adult $m^s \preceq m^t$, Bank $m^s \text{ ? } m^t$, 
and Thyroid $m^s \text{ ? } m^t$). 
On the other hand, for the MLP model class, 
the non-parametric adjustment outperforms 
all source predictors in the semi-synthetic datasets. 
Thus, depending on the model class, non-parametric adjustment 
may not always have a consistent effect on performance.

The generally worse performance of imputation in synthetic and semi-synthetic experiments 
(Table \ref{tab:semisynthetic_perf}) 
helps highlight the difficulty 
of not having missing data indicators. 
Learning without missing data indicators 
is fundamentally more difficult than learning with them, 
and methods which might make sense when missing data indicators are present (e.g. imputation) 
can be ill-defined when the indicators are absent. 
In the eICU dataset, for example, all covariates were binary, 
and so imputing all 0's only left 1's to train on.
As a result, MissForest learned to predict 1 for everything,
rendering these binary features useless. 
Nevertheless, we included a comparison with imputation 
of all zeros in the other datasets, 
as it could still be useful for continuous variables. 

The experiments with real eICU data also help demonstrate 
that it is important to clarify assumptions 
on whether one is truly in a DAMS with UCAR setting,
as failure to do so could result in predictors 
that perform worse than if no adaptation 
had been done in the first place (Table \ref{tab:eicu_results}). 
Ideally, in real-world data, DAMS with UCAR might be useful around a sudden change in clerical practices where the underlying $P(X,Y)$ is similar before and after the change, and underreporting is completely at random 
(e.g. determined based a blanket policy independent of covariates). 
In the absence of such data, however, we instead included synthetic and semisynthetic data where the missingness shift with UCAR assumptions hold, and also included a real critical care (eICU) dataset containing multiple hospitals for thoroughness. 
While our proposed techniques for DAMS with UCAR do not work particularly well on real eICU data, we also note that
we have no particular reason to believe that missingness shift is especially prominent between the hospitals compared to factors such as selection bias (very different cohort), label shift, or changes in prevalences of disease, among others. 
Finding appropriate real world empirical testbeds and analyzing sensitivity to assumption violations are important directions for future work. 

Beyond the UCAR setting, there are several open avenues for further research in domain adaptation under missingness shift. Allowing underreporting to depend on other covariates would significantly broaden the set of applicable real-world cases, as 
doctors often take 
certain measurements as needed in their diagnostic process.
Moreover, future works could explore other 
variations of graphical model structures (Figure \ref{fig:dgp}) 
for expressing 
models of missingness shift.

%% file: sections/80_appendix.tex
\section{Motivating Examples}\label{app:motivating_examples}
\input{appendix/A10_motivating_examples}

\clearpage
\section{DAMS with Indicators as an Instance of Covariate Shift}\label{app:covariate_shift}
\input{appendix/A20_indicators_covariate_shift}

\section{Constant Missingness as L2 Regularization}\label{sec:l2_reg_approx}
\input{appendix/A30_l2_regularization}

\section{Identification of Clean Distribution from Corrupted Distribution}\label{app:clean_from_corrupted}
\input{appendix/A40_id_clean_from_corrupted}

\clearpage
\section{Identification of Labeled Target Distribution from the Labeled Source Distribution}\label{app:id_joint_from_source}
\input{appendix/A50_id_target_from_source}

\section{Error Bound for Estimating Non-Missing Proportions}
\label{sec:nonmissing_prop_samples}
\input{appendix/A60_hoeffdings}

\newpage
\section{Justification for the Non-parametric Procedure with Non-Negative Relative Missingness}\label{app:alternate_nonparametric_pf}
\input{appendix/A70_nonparametric_justification}

\newpage
\section{Optimal Linear Predictors}
\label{app:any_target_missingness}
\input{appendix/A80_linear_predictors}

\newpage

\section{Experiment Details}
\label{app:experiment_details}
\input{appendix/A90_experiment_details}

%% file: appendix/A10_motivating_examples.tex
\textbf{Example \ref{example:redundant_features} (Redundant Features)}

Let $m_s = [1-\epsilon, \epsilon]$ and $m_t = [\epsilon, 1-\epsilon]$. Consider the following data generating process:
\begin{equation*}
\begin{aligned}[c]
Z &= u_Z\\
X_1 &= Z \\
X_2 &= Z \\
Y &= Z + u_Y
\end{aligned}
\qquad \qquad
\begin{aligned}[c]
u_Z &\sim \mathcal{N}(0, \sigma^2_Z)\\
u_Y &\sim \mathcal{N}(0, \sigma^2_Y)
\end{aligned}
\end{equation*}

where $Z$ is a latent variable, $X_1$ and $X_2$ are observed covariates, and $Y$ is the label we wish to predict. 

We start by summarizing the findings, and then provide the full algebraic justification. The optimal (risk-minimizing) linear predictor on the source data is given by:
$$\beta^s_* = \left [\frac{\epsilon}{1 - \epsilon + \epsilon^2}, \frac{1 - \epsilon}{1 - \epsilon + \epsilon^2}\right ]$$
And for the target data:
$$\beta^t_* = \left [\frac{1 - \epsilon}{1 - \epsilon + \epsilon^2}, \frac{\epsilon}{1 - \epsilon + \epsilon^2}\right ]$$

The excess risk of the source predictor on the target data is given by:
\begin{align*}
    r^t(\beta^s_*) - r^t(\beta^t_*) 
    &= (\beta^s_* - \beta^t_*)^\top \smallE{\widetilde{X}^{t\top} \widetilde{X}^t} (\beta^s_* - \beta^t_*)\\ 
    &= \sigma_Z^2 \cdot \frac{(1 - 2\epsilon)^2(1 - 2\epsilon + 2\epsilon^2)}{(1 - \epsilon + \epsilon^2)^2}
\end{align*}

As $\epsilon \rightarrow 0$, we have: 
\begin{align*}
    \beta^s_* &\rightarrow [0, 1]\\
    \beta^t_* &\rightarrow [1, 0]\\
    r^t(\beta^s_*) - r^t(\beta^t_*) &\rightarrow \sigma_Z^2\\
    r^t([0,0]) - r^t(\beta^t_*) &\rightarrow \sigma_Z^2
\end{align*}
that is, the source classifier performs no better than simply predicting 0 (the mean of $Y$). Thus, $r^t(\beta^s_*) \rightarrow \sigma_Z^2 + \sigma_Y^2 = \text{Var}(Y)$

\begin{proof}
    In the example, we have:
\begin{align*}
    \mathbb{E}[X^T X]
    &= \begin{bmatrix}
        \sigma_Z^2 & \sigma_Z^2\\
        \sigma_Z^2 & \sigma_Z^2\\
        \end{bmatrix}\\
    \mathbb{E}[X^T Y]
    &= \begin{bmatrix}
        \sigma_Z^2\\
        \sigma_Z^2\\
        \end{bmatrix}
\end{align*}
We apply the expressions for $\mathbb{E}[\widetilde{X}^T \widetilde{X}]$
and $\mathbb{E}[\widetilde{X}^\top Y]$ derived in Appendix \ref{app:any_target_missingness}:
\begin{align*}
    \mathbb{E}[\widetilde{X}^\top \widetilde{X}] 
    &= (1-m)(1-m)^\top \odot \E{X^\top X} + \diag{m(1-m^\top)}\diag{\E{X^\top X}}\\
    &= \begin{bmatrix}
        1 - m_1     & (1 - m_1)(1 - m_2) \\
        (1 - m_1)(1 - m_2)              & 1 - m_2\\
        \end{bmatrix} \odot \E{X^\top X}\\
    \mathbb{E}[\widetilde{X}^\top Y]
    &= (1 - m) \odot \mathbb{E}[X^\top Y]
\end{align*}
to get:
\begin{align*}
    \mathbb{E}[\widetilde{X}^{t\top} \widetilde{X}^t]  %
    &= \begin{bmatrix}
        1 - \epsilon                & \epsilon (1 - \epsilon) \\
        \epsilon (1 - \epsilon)     & \epsilon\\
        \end{bmatrix} \cdot \sigma_Z^2\\
    \mathbb{E}[\widetilde{X}^{t\top} \widetilde{X}^t]^{-1}  %
    &= \frac{1}{\sigma_Z^2 \epsilon (1 - \epsilon) (1 - \epsilon + \epsilon^2)} \begin{bmatrix}
        \epsilon & -\epsilon(1 - \epsilon)\\
        -\epsilon(1 - \epsilon) & 1 - \epsilon\\
        \end{bmatrix}\\
    \mathbb{E}[\widetilde{X}^{t \top} Y]  %
    &= \sigma_Z^2 \begin{bmatrix}
        1 - \epsilon \\
        \epsilon\\
    \end{bmatrix}\\
    \beta_*^t &= \mathbb{E}[\widetilde{X}^{t\top} \widetilde{X}^t]^{-1} \mathbb{E}[\widetilde{X}^{t \top} Y]\\
    &= \frac{1}{\epsilon (1 - \epsilon) (1 - \epsilon + \epsilon^2)} \begin{bmatrix}
        \epsilon (1 - \epsilon) + -\epsilon^2 (1 - \epsilon)\\
        -\epsilon (1 - \epsilon)^2 + \epsilon (1 - \epsilon)\\
        \end{bmatrix}\\
    &= \frac{1}{\epsilon (1 - \epsilon) (1 - \epsilon + \epsilon^2)} \begin{bmatrix}
        \epsilon (1 - \epsilon) (1 - \epsilon)\\
        \epsilon (1 - \epsilon) (-(1 - \epsilon) + 1)\\
        \end{bmatrix}\\
    &= \frac{1}{1 - \epsilon + \epsilon^2} \begin{bmatrix}
        1 - \epsilon\\
        \epsilon\\
        \end{bmatrix}.
\end{align*}
Similarly, 
\begin{align*}
    \beta_*^s &= \frac{1}{1 - \epsilon + \epsilon^2} \begin{bmatrix}
        \epsilon\\
        1 - \epsilon\\
        \end{bmatrix},
\end{align*}
so we can compute
\begin{align*}
    \beta^s_* - \beta^t_* 
    &= \frac{1}{1 - \epsilon + \epsilon^2} \begin{bmatrix}
        2\epsilon - 1\\
        - 2\epsilon + 1\\
        \end{bmatrix}\\
    &= \frac{1 - 2\epsilon}{1 - \epsilon + \epsilon^2} \begin{bmatrix}
        - 1\\
        1\\
        \end{bmatrix}
\end{align*}

Now, excess risk is computed as follows:
\begin{align*}
    r^t(\beta^s_*) - r^t(\beta^t_*) 
    &= (\beta^s_* - \beta^t_*)^\top \smallE{\widetilde{X}^{t\top} \widetilde{X}^t} (\beta^s_* - \beta^t_*)\\ 
    &= \frac{(1 - 2\epsilon)^2}{(1 - \epsilon + \epsilon^2)^2} 
    \begin{bmatrix}
        - 1\\
        1\\
    \end{bmatrix}^\top
    \begin{bmatrix}
        1 - \epsilon                & \epsilon (1 - \epsilon) \\
        \epsilon (1 - \epsilon)     & \epsilon\\
        \end{bmatrix} \cdot \sigma_Z^2 \cdot
    \begin{bmatrix}
        - 1\\
        1\\
        \end{bmatrix}\\
    &= \frac{\sigma_Z^2 (1 - 2\epsilon)^2 (1 - 2\epsilon + 2\epsilon^2)}{(1 - \epsilon + \epsilon^2)^2} 
\end{align*}

As $\epsilon \rightarrow 0$, we can see that $r^t(\beta^s_*) - r^t(\beta^t_*) \rightarrow \sigma_Z^2$.

Additionally, we can compute the excess risk of the constant zero classifier:
\begin{align*}
    r^t([0, 0]) - r^t(\beta^t_*) 
    &= \beta^{t\top}_* \smallE{\widetilde{X}^{t\top} \widetilde{X}^t} \beta^t_*\\ 
    &= \frac{1}{(1 - \epsilon + \epsilon^2)^2} \begin{bmatrix}
    1 - \epsilon\\
    \epsilon\\
    \end{bmatrix}^\top
    \begin{bmatrix}
        1 - \epsilon                & \epsilon (1 - \epsilon) \\
        \epsilon (1 - \epsilon)     & \epsilon\\
        \end{bmatrix} \cdot \sigma_Z^2 \cdot
    \begin{bmatrix}
        1 - \epsilon\\
        \epsilon\\
        \end{bmatrix}\\
    &= \frac{\sigma_Z^2}{(1 - \epsilon + \epsilon^2)^2} \begin{bmatrix}
        (1 - \epsilon)^2 + \epsilon^2 (1 - \epsilon)\\
        \epsilon(1 - \epsilon)^2 + \epsilon^2 \\
        \end{bmatrix}^\top
        \begin{bmatrix}
            1 - \epsilon\\
            \epsilon\\
        \end{bmatrix}\\
    &= \frac{\sigma_Z^2 (1 - \epsilon + \epsilon^2) }{(1 - \epsilon + \epsilon^2)^2} \begin{bmatrix}
        (1 - \epsilon)\\
        \epsilon \\
        \end{bmatrix}^\top
        \begin{bmatrix}
            1 - \epsilon\\
            \epsilon\\
        \end{bmatrix}\\
    &= \frac{\sigma_Z^2 (1 - \epsilon + \epsilon^2) }{(1 - \epsilon + \epsilon^2)^2} 
        \left [(1 - \epsilon)^2 + \epsilon^2 \right ]\\
    &= \frac{\sigma_Z^2 (1 - 2\epsilon + 2\epsilon^2)}{1 - \epsilon + \epsilon^2}
\end{align*}
As $\epsilon \rightarrow 0$, we can see that $r^t([0, 0]) - r^t(\beta^t_*) \rightarrow \sigma_Z^2$.
\end{proof}

\clearpage

\textbf{Example \ref{example:confounded} (Confounded Features)}

Now, suppose that $m_s = [0, 0]$ and $m_t = [1, 0]$. For some constants $a, b, c$ consider the following data generating process:
\begin{equation*}
\begin{aligned}[c]
X_1 &= \nu_{1}\\
X_2 &= aX_1 + \nu_{2}\\
Y &= bX_1 + cX_2 + \nu_Y
\end{aligned}
\qquad \qquad
\begin{aligned}[c]
\nu_{1} &\sim \mathcal{N}(0, 1)\\
\nu_{2} &\sim \mathcal{N}(0, 1)\\
\nu_Y &\sim \mathcal{N}(0, 1).
\end{aligned}
\end{equation*}
We will show that the optimal source and target predictors are $\beta^s_* = [b, c]$ and $\beta^t_* = [0, \frac{ab}{a^2 + 1} + c]$. By setting $a = -\frac{b}{c}$, we will show that
for any $\tau > 1$, 
there exists values of $a, b, c$ such that $r^t(\beta_*^s) > \tau \text{Var}(Y)$.

\begin{proof}
First, we compute $\beta_*^s$ (where $m_s = [0, 0]$):
\begin{align*}
    \mathbb{E}[\widetilde{X}^{s \top} \widetilde{X}^s] %
    &= \E{X^\top X}\\
    &= \begin{bmatrix}
        1     & a \\
        a     & a^2 + 1\\
        \end{bmatrix} \\
    \mathbb{E}[\widetilde{X}^{s \top} \widetilde{X}^s]^{-1} %
    &= \begin{bmatrix}
        a^2 + 1          & -a \\
        -a & 1\\
        \end{bmatrix} \\
    \mathbb{E}[\widetilde{X}^{s\top} Y] %
    &= \mathbb{E}[X^\top Y]\\
    &= \begin{bmatrix}
        b + ac \\
        ab + a^2 c + c\\
        \end{bmatrix} \\
    \beta_*^s %
    &= \mathbb{E}[\widetilde{X}^{s \top} \widetilde{X}^s]^{-1} \mathbb{E}[\widetilde{X}^{s\top} Y]\\
    &= \begin{bmatrix}
        a^2 + 1          & -a \\
        -a & 1\\
        \end{bmatrix} 
        \cdot \begin{bmatrix}
        b + ac \\
        ab + a^2 c + c\\
        \end{bmatrix} \\
    &= \begin{bmatrix}
        b \\
        c 
        \end{bmatrix}.
\end{align*}
Thus, $\beta_*^s = [b, c]$.

Now, let us compute $\beta_*^t$ (where $m_t = [1, 0]$). Since $X_1$ is entirely missing and $X_2$ is completely observed, we only regress on $X_2$:
\begin{align*}
    \mathbb{E}[\widetilde{X}_2^{t \top} \widetilde{X}_2^t] %
    &= \mathbb{E}[X_2^\top X_2]\\
    &= (a^2 + 1)\\
    \mathbb{E}[\widetilde{X}_2^{t \top} \widetilde{X}_2^t]^{-1} %
    &= \frac{1}{a^2 + 1}\\
    \mathbb{E}[\widetilde{X}_2^{t\top} Y] %
    &= ab + a^2c + c\\
    \mathbb{E}[\widetilde{X}_2^{t \top} \widetilde{X}_2^t]^{-1} %
    \mathbb{E}[\widetilde{X}_2^{t\top} Y]
    &= \frac{ab + a^2 c + c}{a^2 + 1}\\
    &= \frac{ab}{a^2 +1} + c.
\end{align*}

Thus, $\beta_*^t = \left[ 0 , \frac{ab}{a^2 +1} + c \right]$.

Now, let us compute $\text{Var}(Y)$. Note that $\mathbb{E}[Y] = 0$, so $\text{Var}(Y) = \mathbb{E}[Y^2]$. Also, note that $\nu_1, \nu_2, \nu_Y$ are independent:
\begin{align*}
    \text{Var}(Y) 
    &= \text{Var}(bX_1 + cX_2 + \nu_Y)\\
    &= \mathbb{E}[(b\nu_1 + c(a\nu_1 + \nu_2) + \nu_Y)^2]\\
    &= (b + ac)^2 + c^2 + 1\\
    &= b^2 + 2abc + a^2 c^2 + c^2 + 1.
\end{align*}

Thus, $\text{Var}(Y) = b^2 + 2abc + a^2 c^2 + c^2 + 1$.

Now, let us compute $r^t(\beta_*^s)$. Let $[\beta_*^s]_2$ denote the second dimension of $\beta_*^s$. We have:
\begin{align*}
    r^t(\beta_*^s)
    &= \mathbb{E}[(Y - \widetilde{X}^t_2 [\beta_*^s]_2)^2]\\
    &= \mathbb{E}[Y^2] - 2\mathbb{E}[\widetilde{X}^t_2 [\beta_*^s]_2 Y] + \mathbb{E}[(\widetilde{X}^t_2 [\beta_*^s]_2)^2]\\
    &= \text{Var}[Y^2] - 2\mathbb{E}[X_2 [\beta_*^s]_2 Y] + \mathbb{E}[(X_2 [\beta_*^s]_2)^2]\\
    &= \text{Var}[Y^2] 
        - 2\mathbb{E}[(a\nu_1 + \nu_2) c (b\nu_1 + c(a\nu_1 + \nu_2) + \nu_Y)] 
        + \mathbb{E}[((a\nu_1 + \nu_2)c)^2]\\
    &= b^2 + 2abc + a^2 c^2 + c^2 + 1 - 2[ac(b+ac) + c^2] + [a^2 c^2 + c^2]\\
    &= b^2 + 1.
\end{align*}

Thus, we have $\frac{r^t(\beta_*^s)}{\text{Var}(Y)} = \frac{b^2 + 1}{b^2 + 2abc + a^2 c^2 + c^2 + 1}$. If we set $a = -\frac{b}{c}$, then we have:
\begin{align*}
\frac{r^t(\beta_*^s)}{\text{Var}(Y)} 
&= \frac{b^2 + 1}{b^2 + 2abc + a^2 c^2 + c^2 + 1}\\
&= \frac{b^2 + 1}{b^2 - 2b^2 + b^2 + c^2 + 1}\\
&= \frac{b^2 + 1}{c^2 + 1}.
\end{align*}

Now suppose that for some $\tau > 1$, we would like $r^t(\beta_*^s) > \tau \text{Var}(Y)$. Then, it is easy to see that we can simply choose $b$ large enough, $c$ small enough, and $a = -\frac{b}{c}$, such that $\frac{b^2 + 1}{c^2 + 1} > \tau$.
\end{proof}

%% file: appendix/A20_indicators_covariate_shift.tex
This section contains a proof of Proposition \ref{prop:covshift}: 
Assume we observe $\xi$. Let us consider an augmented set of covariates $\tilde{x}' = (\tilde{x}, \xi)$. 
When $\xi$ is drawn independently of other covariates or depending only on other completely observed covariates, we will show that missingness shift satisfies the covariate shift assumption,
i.e, $P^s(Y|\widetilde{X}' = \tilde{x}') = P^t(Y|\widetilde{X}' = \tilde{x}')$.

First, let us formalize what it means for $\xi$ to be drawn independently of other covariates or depending only on other completely observed covariates:

\begin{enumerate}[(a)]
    \item \textbf{Independent of other covariates} When $\xi$ is drawn independently of other covariates, as described in the DAMS with UCAR setup (Section \ref{sec:missshift_definition}), we have that $\xi \sim \text{Bernoulli}(1 - m)$ for some constant vector of missingness rates $m \in [0, 1]^d$.
    \item \textbf{Depending only on other completely observed covariates} Now, suppose that some subset of covariates $X_c \subseteq X$ is completely observed (i.e. no missingness), and the missingness of the other covariates $X_m = X \setminus X_c$ depends on $X_c$. That is, $\xi \sim \text{Bernoulli}(f(X_c))$ for some function $f: \mathbb{R}^{|X_c|} \rightarrow [0, 1]^{|X_m|}$.
\end{enumerate}

Since (b) is more general than (a), we adopt notation from (b) throughout our proof, and then argue why it also holds for (a).

\begin{proof}
Consider some augmented set of covariates taking values $\tilde{x}' = (\tilde{x}_m, \xi, x_c)$.
To prove that the covariate shift assumption holds, let us start by considering the left-hand side of the equation. Applying Bayes' Rule, we have:
\begin{align*}
     P^s(Y|\widetilde{X}' = \widetilde{x}')
     &= P^s(Y|\widetilde{X}^s_m = \widetilde{x}_m, \xi^s = \xi, X_c = x_c) = \frac{P^s(Y, \widetilde{X}^s_m = \widetilde{x}_m, \xi^s = \xi, X_c = x_c)}{\sum_{y} P^s(Y = y, \widetilde{X}^s_m = \widetilde{x}_m, \xi^s = \xi, X_c = x_c)}
\end{align*}

We can rewrite the numerator as follows:
\begin{align*}
    P^s(Y, \widetilde{X}^s_m = \widetilde{x}_m, \xi^s = \xi, X_c = x_c)
    &= \sum_{x_m: x_m \odot \xi = \widetilde{x}_m} P(Y, \widetilde{X}^s_m = \widetilde{x}_m, \xi^s = \xi, X_m = x_m, X_c = x_c)\\
    &= \sum_{x_m: x_m \odot \xi = \widetilde{x}_m} P(Y, \xi^s = \xi, X_m = x_m, X_c = x_c)\\
    &= \sum_{x_m: x_m \odot \xi = \widetilde{x}_m} P(\xi^s = \xi | Y, X_m = x_m, X_c = x_c) \cdot P(Y, X_m = x_m, X_c = x_c)\\
    &= \sum_{x_m: x_m \odot \xi = \widetilde{x}_m} P(\xi^s = \xi | X_c = x_c) \cdot P(Y, X_m = x_m, X_c = x_c)\\
    &= P(\xi^s = \xi | X_c = x_c)  \sum_{x_m: x_m \odot \xi = \widetilde{x}_m} P(Y, X_m = x_m, X_c = x_c),
\end{align*}
where the first line follows from marginalizing over all possible values of $X_m$, the second line comes from the fact that $\widetilde{x}_m$ is determined given $x_m$ and $\xi$, the third line comes from Bayes' Rule, the fourth line comes the fact that $\xi$ only depends on $X_c$, and the last line comes from pulling the first term out of the summation.

Plugging back into the expression for $P^s(Y|\widetilde{X}' = \widetilde{x}')$, we have:
\begin{align*}
    P^s(Y|\widetilde{X}' = \widetilde{x}') 
    &= \frac{P^s(Y, \widetilde{X}^s_m = \widetilde{x}_m, \xi^s = \xi, X_c = x_c)}{\sum_{y} P^s(Y = y, \widetilde{X}^s_m = \widetilde{x}_m, \xi^s = \xi, X_c = x_c)}\\
    &= \frac{P(\xi^s = \xi | X_c = x_c)  \sum_{x_m: x_m \odot \xi = \widetilde{x}_m} P(Y, X_m = x_m, X_c = x_c)}{\sum_y P(\xi^s = \xi | X_c = x_c)  \sum_{x_m: x_m \odot \xi = \widetilde{x}_m} P(Y=y, X_m = x_m, X_c = x_c)}\\
    &= \frac{\sum_{x_m: x_m \odot \xi = \widetilde{x}_m} P(Y, X_m = x_m, X_c = x_c)}{\sum_{x_m: x_m \odot \xi = \widetilde{x}_m} P(Y=y, X_m = x_m, X_c = x_c)},
\end{align*}

which does not contain source-specific quantities (everything is in terms of the underlying distribution). By the same logic, 
$$P^t(Y|\widetilde{X}' = \widetilde{x}') = \frac{\sum_{x_m: x_m \odot \xi = \widetilde{x}_m} P(Y, X_m = x_m, X_c = x_c)}{\sum_{x_m: x_m \odot \xi = \widetilde{x}_m} P(Y=y, X_m = x_m, X_c = x_c)}.$$

Thus, $P^s(Y|\widetilde{X}' = \widetilde{x}') = P^t(Y|\widetilde{X}' = \widetilde{x}')$ as desired. 
When $\xi$ is instead drawn independently of other covariates, as in (a) above, we note that all of the steps of the proof follow through simply by removing $X_c$.
Additionally, while all of the above expressions apply to discrete $X$, extension to continuous $X$ is straightforward (e.g. replace summations with integrals, and constants with sets or intervals).
\end{proof}

%% file: appendix/A30_l2_regularization.tex
This section contains a proof of Theorem \ref{theorem:l2_reg_approx}. This proof is based off of that presented in \citet{NIPS2013_38db3aed_dropout}'s work showing dropout to be a form of adaptive regularization. Instead of assuming a single constant dropout rate across all covariates, however, our proof extends to varying rates of missingness (i.e. different constant dropout rates) for different covariates. 

\begin{proof}
Assume we know the constant missingness rates $m$. For mathematical convenience, we preprocess $\widetilde{x}$ by multiplying each dimension by the corresponding $\frac{1}{1-m_j}$. For the remainder of this derivation, this preprocessed data is referred to as $\widetilde{x}$.

Similar to \citet{NIPS2013_38db3aed_dropout}, we start with an analysis of generalized linear models and then consider the case of linear regression. Minimizing the expected negative log likelihood $l_{\widetilde{x}^{(i)}, y^{(i)}}(\beta)$ of a generalized linear model $p_\beta(y|x) = h(y)\exp\{yx\cdot \beta - A(x \cdot \beta)\}$, we have:
\begin{align*}
    \widehat{\beta} &= \arg\min_{\beta \in \mathbb{R}^d} \sum_{i=1}^n \mathbb{E}_{\xi}[l_{\widetilde{x}^{(i)}, y^{(i)}}(\beta)]\\
    \sum_{i=1}^n \mathbb{E}_{\xi}[l_{\widetilde{x}^{(i)}, y^{(i)}}(\beta)] 
    &= \sum_{i=1}^n \mathbb{E}_{\xi}[-\log p_{\beta}(y^{(i)} | \widetilde{x}^{(i)})]\\
    &= \sum_{i=1}^n \mathbb{E}_{\xi}[-(\log h(y^{(i)}) + y^{(i)} \widetilde{x}^{(i)} \beta - A(\widetilde{x}^{(i)} \cdot \beta))]\\
    &= \sum_{i=1}^n -\log h(y^{(i)}) - y^{(i)} \mathbb{E}_{\xi}[\widetilde{x}^{(i)}] \beta + \mathbb{E}_{\xi}[A(\widetilde{x}^{(i)} \cdot \beta)]\\
    &= \sum_{i=1}^n -\log h(y^{(i)}) - y^{(i)}\left (x^{(i)} \odot \frac{1 - m}{1 - m} \right ) \beta + \mathbb{E}_{\xi}[A(\widetilde{x}^{(i)} \cdot \beta)]\\
    &= \sum_{i=1}^n -(\log h(y^{(i)}) + y^{(i)} x^{(i)} \beta - A(x^{(i)} \beta)) - A(x^{(i)} \beta) + \mathbb{E}_{\xi}[A(\widetilde{x}^{(i)} \cdot \beta)]\\
    &= \sum_{i=1}^n l_{x^{(i)}, y^{(i)}}(\beta) + \mathbb{E}_{\xi}[A(\widetilde{x}^{(i)} \cdot \beta)] - A(x^{(i)} \beta)\\
    &= \sum_{i=1}^n l_{x^{(i)}, y^{(i)}}(\beta) + R(\beta)
\end{align*}

where $R(\beta) \triangleq \sum_{i=1}^{n}\mathbb{E}_{\xi}[A(\widetilde{x}^{(i)} \cdot \beta)] - A(x^{(i)} \beta)$. How do we interpret $R(\beta)$?

First, we do a second order Taylor expansion of $A$ around $x\beta$. Note that linear regression has a second order log partition function. Thus, for linear regression this expansion is exact:
    \begin{align*}
        A(y) &\approx A(x\beta) + A'(x\beta)(y-x\beta) + \frac{1}{2}A''(x\beta)(y-x\beta)^2\\
        A(\widetilde{x}\beta) &\approx A(x\beta) + A'(x\beta)(\widetilde{x}\beta - x\beta) + \frac{1}{2}A''(x\beta)(\widetilde{x}\beta - x\beta)^2\\
        &= A(x\beta) + A'(x\beta)(\widetilde{x} - x)\beta + \frac{1}{2}A''(x\beta)(\widetilde{x}\beta - x\beta)^2
    \end{align*}
Now, we can compute the first term of $R(\beta)$:
    \begin{align*}
        \mathbb{E}_{\xi}[A(\widetilde{x} \cdot \beta)] 
        &\approx  \mathbb{E}_{\xi}[A(x \beta)] + \mathbb{E}_{\xi}[A'(x \beta)(\widetilde{x} - x)\beta] + \mathbb{E}_{\xi}[\frac{1}{2}A''(x\beta)(\widetilde{x}\beta - x\beta)^2]\\
        &= A(x\beta) + 0 + \frac{1}{2}A''(x\beta)\mathbb{E}_\xi [(\widetilde{x}\beta - x\beta)^2]\\
        &= A(x\beta) + \frac{1}{2}A''(x\beta)\text{Var}_\xi (\widetilde{x}\beta)
    \end{align*}
    where the second step follows because $\mathbb{E}_{\xi}[\widetilde{x}] = x$.
Thus, $R(\beta)$ is given by:
    \begin{align*}
        R(\beta) &= \sum_{i=1}^n \mathbb{E}_{\xi}[A(\widetilde{x}^{(i)} \cdot \beta)] - A(x^{(i)} \beta)\\
        &\approx \sum_{i=1}^n A(x^{(i)}\beta) + \frac{1}{2}A''(x^{(i)}\beta)\text{Var}_\xi (\widetilde{x}^{(i)}\beta) - A(x^{(i)} \beta)\\
        &= \sum_{i=1}^n \frac{1}{2}A''(x^{(i)}\beta)\text{Var}_\xi (\widetilde{x}^{(i)}\beta)\\
        &\triangleq R^q(\beta).
    \end{align*}
    Note that the first term corresponds to variance of $y^{(i)}$, and the second term corresponds to the variance of the estimated GLM parameter due to noising, or in the linear case, $\text{Var}(y^{(i)})$. Additionally, note that for linear regression $R(\beta) = R^q(\beta)$ since the approximate equality comes from the Taylor series approximation.
    
Analyzing $\text{Var}_\xi (\widetilde{x}^{(i)}\beta)$,
    \begin{align*}
        \text{Var}_\xi (\widetilde{x}^{(i)}\beta) 
        &= \sum_{j=1}^d \text{Var}_\xi (\widetilde{x}_{j}^{(i)}\beta_j)\\
        &= \sum_{j=1}^d \text{Var}_\xi \left ( \frac{x_{j}^{(i)}}{1 - m_j} \cdot b_j \cdot \beta_j \right )\\
        &= \sum_{j=1}^d \left ( \frac{x_{j}^{(i)}}{1 - m_j}\right )^2 \beta_j^2 (1 - m_j)(m_j)\\
        &= \sum_{j=1}^d \frac{m_j}{1 - m_j} \left (x_{j}^{(i)} \right)^2 \beta_j^2
    \end{align*}
    where $b_j \sim \text{Bernoulli}(1 - m_j)$.
Thus, $R^q(\beta)$ is given by:
    \begin{align*}
        R^q(\beta) &= \frac{1}{2} \sum_{i=1}^{n} A''(x^{(i)}\beta) \sum_{j=1}^d \frac{m_j}{1 - m_j} \left(x_{j}^{(i)}\right)^2 \beta_j^2.
    \end{align*}
    Let $V(\beta) \in \mathbb{R}^{n\times n}$ be diagonal with entries $A''(x^{(i)}\beta)$, and $X \in \mathbb{R}^{n\times d}$ be the design matrix with rows $x^{(i)}$. For linear regression, $V(\beta)$ is given by the identity matrix. Then, we can rewrite $R^q(\beta)$ as:
    \begin{align*}
        R^q(\beta) &= \frac{1}{2} \left (\beta \odot \sqrt{\frac{m}{1-m}}\right )^\top \text{diag}(X^\top V(\beta) X) \left (\beta \odot \sqrt{\frac{m}{1-m}}\right )\\
        R^q(\beta)
        &= \frac{1}{2} \left (\beta \odot \frac{m}{1-m} \right )^\top \text{diag}(I) \left (\beta \odot \frac{m}{1-m}\right )\\
        &= \frac{1}{2} \left (\text{diag}(I)^{1/2} \beta \odot \frac{m}{1-m} \right )^\top \left (\text{diag}(I)^{1/2} \beta \odot \frac{m}{1-m}\right )\\
        &= \frac{1}{2} \left (\beta \widetilde{\Delta}_{\text{diag}} \right )^\top \left (\beta \widetilde{\Delta}_{\text{diag}} \right )
    \end{align*}
    where $\widetilde{\Delta}_{\text{diag}} = \text{diag}\left ( \sqrt{\frac{m}{1 - m}} \right ) \text{diag}(I)^{1/2}$, where $\text{diag}\left(\sqrt{\frac{m}{1-m}}\right)$ refers to a diagonal matrix with the vector quantities on the diagonal, and $\text{diag}(I)^{1/2}$ refers to the square root of the diagonal of the Fisher information matrix. Thus, for linear regression, applying missingness rates $m \in [0, 1]^d$ to data scaled by $\frac{1}{1-m}$ can be viewed as an attempt to apply L2 regularization of $\beta$ scaled by  $\widetilde{\Delta}_{\text{diag}}$.
\end{proof}

%% file: appendix/A40_id_clean_from_corrupted.tex
This section proves Lemma \ref{lemma:clean_from_corrupted}, which states that the clean distribution $p$ is identified from the corrupted distribution $\widetilde{p}$ given missingness rates $m$, and $m\prec 1$.

\begin{proof}
Let $\mathcal{A}^k$ denote the set of possible values of $x$ where at most $k$ of the dimensions of $x$ are 0. We would like to show that $\forall k \in \{0, 1, ..., d\}$, $\forall a \in \mathcal{A}^k$, the clean distribution $p_{a,y}$ is identifiable (and hence $p_{x,y}$ is identifiable) for all values of $x$ and $y$. We proceed with a proof by induction on $k$.

\begin{itemize}
    \item \emph{Base case ($k=0$)}: 
    
    Consider $\mathcal{A}^0$, the set of possible values of $x$ where none of the dimensions of $x$ are 0. For any subset $a \subseteq \mathcal{A}^0$, we can write:
    $$\widetilde{p}_{a,y} = \prod_{j=1}^d (1-m_j) p_{a,y}$$
    which can be rearranged to recover $p_a$ from $\widetilde{p}_a$ and $m$, which are both known:
    $$p_{a,y} = \prod_{j=1}^d \frac{1}{1-m_j}\widetilde{p}_{a,y}.$$
    Thus $p_{a,y}$ is identified for $a\subseteq \mathcal{A}^0$.

    \item \emph{Inductive Step}: 
    Assume $p_{a,y}$ is identified for $a \subseteq \mathcal{A}^{k}$. Consider some $a' \subseteq \mathcal{A}^{k+1}$. Using equation $\eqref{eqn:elementwise_corruption}$, we have:
    \begin{align*}
        \widetilde{p}_{a',y} &= \sum_{b: b \leadsto a'} p_{b, y} \cdot \prod_{j=1}^d (1-m_{j})^{[a'_j]_{\neq 0}} m_{j}^{[b_j]_{\neq 0} - [a'_j]_{\neq 0}}\\
        &= p_{a', y} \cdot \prod_{j=1}^d (1-m_{j})^{[a'_j]_{\neq 0}} + \sum_{\substack{b: b \leadsto a',\\ b \neq a'}} p_{b, y} \cdot \prod_{j=1}^d (1-m_{j})^{[a'_j]_{\neq 0}} m_{j}^{[b_j]_{\neq 0} - [a'_j]_{\neq 0}}
    \end{align*}
    
    Recall from Remark \ref{remark:mreach_characters} that if $b \leadsto a'$, then the dimensions of $b$ that are 0 must be a subset of the ones that are 0 in $a'$. Additionally, any dimensions that are nonzero in both $b$ and $a'$ must match in value. This implies that if there are the same number of zeros in $b$ and $a'$, then $b = a'$. The remaining $b$ where $b\leadsto a'$ have at least one less zero than $a'$. Thus, the set of $\{b: b\leadsto a', b\neq a'\} \in \mathcal{A}^k$, and by our inductive hypothesis, $p_{b, y}$ are identified when $b \in \mathcal{A}^k$. As a result, we can identify the second term in the equation above (the summation over $b$'s), and rearranging the equation, we can identify $p_{a', y}$ as $\widetilde{p}$ and $m$ are known.
\end{itemize}
Thus, by the principle of mathematical induction, $p_a$ is identified for $a\in\mathcal{A}^k$, $\forall k \in \{0, 1, ..., d\}$. Therefore, given $m$, we have identified the clean distribution from the corrupted distribution. Additionally, while all of the above expressions apply to discrete $X$, extension to continuous $X$ is straightforward (e.g. replace summations with integrals, and constants with sets or intervals).
\end{proof}

%% file: appendix/A50_id_target_from_source.tex
Here we prove Theorem \ref{theorem:s_to_t_elementwise}, which states that:
$$\widetilde{p}^t_{x,y} = \sum_{z: z \leadsto x} \widetilde{p}^s_{z, y} \cdot \prod_{j=1}^d (1-r_{j}^{s \rightarrow t})^{[x_j]_{\neq 0}} (r_{j}^{s \rightarrow t})^{[z_j]_{\neq 0} - [x_j]_{\neq 0}}$$

\begin{proof}
Applying equation \eqref{eqn:elementwise_corruption}, the corrupted source and target distributions can be written as:
\begin{align*}
    \widetilde{p}^s_{a,y} &= \sum_{b: b \leadsto a} p_{b, y} \cdot \prod_{j=1}^d (1-m_{sj})^{[a_j]_{\neq 0}} m_{sj}^{[b_j]_{\neq 0} - [a_j]_{\neq 0}}\\
    \widetilde{p}^t_{a,y} &= \sum_{c: c \leadsto a} p_{c, y} \cdot \prod_{j=1}^d (1-m_{tj})^{[a_j]_{\neq 0}} m_{tj}^{[c_j]_{\neq 0} - [a_j]_{\neq 0}}
\end{align*}

We apply relative missingness $r = r^{s\rightarrow t} =\frac{m_t - m_s}{1-m_s}$ to source distribution $\widetilde{p}^s$, denoting this new distribution as $\widetilde{p}^{s\rightarrow t}$:
\begin{align*}
    \widetilde{p}_{a,y}^{s\rightarrow t} 
    &= \sum_{b: b \leadsto a} \widetilde{p}^s_{b, y} \cdot \prod_{j=1}^d (1-r_j)^{[a_j]_{\neq 0}} r_{j}^{[b_j]_{\neq 0} - [a_j]_{\neq 0}}\\
    &=\sum_{b: b \leadsto a} \sum_{c: c \leadsto b} p_{c, y} \cdot \prod_{j=1}^d (1-m_{sj})^{[b_j]_{\neq 0}} m_{sj}^{[c_j]_{\neq 0} - [b_j]_{\neq 0}} \cdot \prod_{j=1}^d (1-r_j)^{[a_j]_{\neq 0}} r_{j}^{[b_j]_{\neq 0} - [a_j]_{\neq 0}}\\
    &= \sum_{c: c \leadsto b} p_{c, y} \sum_{b: b \leadsto a} \cdot \prod_{j=1}^d (1-m_{sj})^{[b_j]_{\neq 0}} m_{sj}^{[c_j]_{\neq 0} - [b_j]_{\neq 0}} \cdot \prod_{j=1}^d (1-r_j)^{[a_j]_{\neq 0}} r_{j}^{[b_j]_{\neq 0} - [a_j]_{\neq 0}}\\
    &= \sum_{c: c \leadsto b} p_{c, y} \sum_{b: b \leadsto a} \cdot \prod_{j=1}^d (1-m_{sj})^{[b_j]_{\neq 0}} m_{sj}^{[c_j]_{\neq 0} - [b_j]_{\neq 0}} \cdot \prod_{j=1}^d \left(\frac{1- m_{tj}}{1-m_{sj}}\right)^{[a_j]_{\neq 0}} \left(\frac{m_{tj} - m_{sj}}{1-m_{sj}}\right)^{[b_j]_{\neq 0} - [a_j]_{\neq 0}}\\
    &= \sum_{c: c \leadsto b} p_{c, y} \sum_{b: b \leadsto a} \prod_{j=1}^d 
    (1-m_{sj})^{\mathbbm{1}\left\{[c_j]_{\neq 0} = [b_j]_{\neq 0} = 1, [a_j]_{\neq 0} = 0\right\} + \mathbbm{1}\left\{[c_j]_{\neq 0} = [b_j]_{\neq 0} = [a_j]_{\neq 0} = 1\right\}} \\
        &\hspace{8em}\cdot m_{sj}^{\mathbbm{1}\left\{[c_j]_{\neq 0} = 1, [b_j]_{\neq 0} = [a_j]_{\neq 0} = 0\right\}}\\
        &\hspace{8em}\cdot \left(\frac{1- m_{tj}}{1-m_{sj}}\right)^{\mathbbm{1}\left\{[c_j]_{\neq 0} = [b_j]_{\neq 0} = [a_j]_{\neq 0} = 1\right\}}\\ 
        &\hspace{8em}\cdot \left(\frac{m_{tj} - m_{sj}}{1-m_{sj}}\right)^{\mathbbm{1}\left\{[c_j]_{\neq 0} = [b_j]_{\neq 0} = 1, [a_j]_{\neq 0} = 0\right\}}\\
    &= \sum_{c: c \leadsto b} p_{c, y} \sum_{b: b \leadsto a} \prod_{j=1}^d 
        m_{sj}^{\mathbbm{1}\left\{[c_j]_{\neq 0} = 1, [b_j]_{\neq 0} = [a_j]_{\neq 0} = 0\right\}}\\
        &\hspace{8em}\cdot (1-m_{tj})^{\mathbbm{1}\left\{[c_j]_{\neq 0} = [b_j]_{\neq 0} = [a_j]_{\neq 0} = 1\right\}} \\
        &\hspace{8em}\cdot \left(m_{tj} - m_{sj}\right)^{\mathbbm{1}\left\{[c_j]_{\neq 0} = [b_j]_{\neq 0} = 1, [a_j]_{\neq 0} = 0\right\}}\\
    &= \sum_{c: c \leadsto a} p_{c, y} 
        \cdot \left(\prod_{j: [c_j]_{\neq 0} = [a_j]_{\neq 0} = 1}1-m_{tj}\right) 
        \cdot \left(\prod_{j: [c_j]_{\neq 0} = [a_j]_{\neq 0} = 0} 1 \right)\\
        &\hspace{6em}\cdot \sum_{b: b \leadsto a} \left(\prod_{j: [c_j]_{\neq 0} = 1, [a_j]_{\neq 0} = 0} 
        m_{sj}^{1-[b_j]_{\neq 0}} (m_{tj}-m_{sj})^{[b_j]_{\neq 0}}\right)\\
    &= \sum_{c: c \leadsto a} p_{c, y} 
        \cdot \left(\prod_{j: [c_j]_{\neq 0} = [a_j]_{\neq 0} = 1}1-m_{tj}\right) 
        \cdot \left(\prod_{j: [c_j]_{\neq 0} = [a_j]_{\neq 0} = 0} 1 \right)\\
        &\hspace{6em}\cdot \sum_{[b]_{\neq 0} \in \{0,1\}^d} \left(\prod_{j: [c_j]_{\neq 0} = 1, [a_j]_{\neq 0} = 0} 
        m_{sj}^{1-[b_j]_{\neq 0}} (m_{tj}-m_{sj})^{[b_j]_{\neq 0}}\right)\\
    &= \sum_{c: c \leadsto a} p_{c, y} 
        \cdot \left(\prod_{j: [c_j]_{\neq 0} = [a_j]_{\neq 0} = 1}1-m_{tj}\right) 
        \cdot \left(\prod_{j: [c_j]_{\neq 0} = [a_j]_{\neq 0} = 0} 1 \right) \cdot \left(\prod_{j: [c_j]_{\neq 0} = 1, [a_j]_{\neq 0} = 0} m_{t_j} \right)\\
    &= \sum_{c: c \leadsto a} p_{c, y} 
        \prod_{j=1}^d (1-m_{tj})^{[a_j]_{\neq 0}} m_{tj}^{[c_j]_{\neq 0} - [a_j]_{\neq 0}} \\
    &= \widetilde{p}^t_{a,y}
\end{align*}

as desired. The steps are explained in words below:
\begin{itemize}
    \item Plug in equation for corrupted source distribution.
    \item Switch summation order and factor out $p_{c,y}$.
    \item Plug in for $r$.
    \item Note that $[c_j]_{\neq 0} - [b_j]_{\neq 0} = 1$ only if $[c_j]_{\neq 0} = 1$ and $[b_j]_{\neq 0} = 0$. Use similar reasoning for the remaining, keeping in mind that $[c]_{\neq 0} \succeq [b]_{\neq 0} \succeq [a]_{\neq 0}$. Simplify.
    \item Since all elements of the sum have $\mathbbm{1}\{[c]_{\neq 0} \succeq [b]_{\neq 0} \succeq [a]_{\neq 0}\}$, it is also true that $\mathbbm{1}\{[c]_{\neq 0} \succeq [a]_{\neq 0}\}$.
    \item If $[a_i]_{\neq 0} = [c_i]_{\neq 0} = 1$, then $[b_i]_{\neq 0} = 1$ necessarily. 
    \item Note that if $c\leadsto b\leadsto a$ and $[c_i]_{\neq0} = 1, [a_i]_{\neq0} =0$, then $\forall i, b_i \in \{0, c_i\}$. We can then perform a change of variables in the summation, now summing over $[b]_{\neq0} \in \{0,1\}^d$ instead.
    \item We use the following identity for arbitrary $d$-dimensional vectors $a$ and $b$:
    \begin{align*}
        \sum_{u \in \{0, 1\}^d}\prod_j a_j^{u_j} b_j^{1 - u_j} &= \prod_j (a_j + b_j)
    \end{align*}
    To gain intuition for why this is the case, let's start with $d = 2$:
    \begin{align*}
        LHS &= \sum_{u \in \{0, 1\}^d}\prod_j a_j^{u_j} b_j^{1 - u_j}\\
        &= \sum_{u \in \{0, 1\}^2} a_1^{u_1} b_1^{1 - u_1} a_2^{u_2} b_2^{(1 - u_2)}\\
        &= \sum_{u \in [(1, 1), (1, 0), (0, 1), (0, 0)]} a_1^{u_1} b_1^{1 - u_1} a_2^{u_2} b_2^{(1 - u_2)}\\
        &=  a_1 a_2 + a_1 b_2 +  b_1 a_2 + b_1 b_2\\
        RHS &= \prod_j (a_j + b_j)\\
        &= (a_1 + b_1)(a_2 + b_2)\\
        &= a_1 a_2 + a_1 b_2 + b_1 a_2 + b_1 b_2
    \end{align*}
    Notice that the right-hand side is a product of sums $(a_j + b_j)$, of which there are $d$ terms. When expanding this product of sums into a sum of products, each term in the sum of products will include either $a_j$ or $b_j$ for all $j \in {1, 2, ..., d}$. Summing over all possible choices of either $a_j$ or $b_j$ for all $j$ is then equivalent to summing over all possible values of a binary $d$-dimensional vector $u$. Thus, we get the left-hand side of the identity.
    \item The remaining steps are straightforward simplifications to get a form matching equation \eqref{eqn:s_to_t_elementwise}.
    \item Note that while all of the above expressions apply to discrete $X$, extension to continuous $X$ is straightforward (e.g. replace summations with integrals, and constants with sets or intervals).
\end{itemize}

\end{proof}

%% file: appendix/A60_hoeffdings.tex
This is a proof of Theorem \ref{theorem:relative_miss_estimation}.
To estimate the non-missingness proportion $q = P(\widetilde{X} = 1)$ within $\epsilon$ of the true non-missingness proportion with probability at least $1 - \delta$, we use Hoeffding's bound to show:
\begin{align*}
    P(|\widehat{q} - q| \geq \epsilon) &\leq 2\exp(-2ne^2) = \delta\\
    \implies -2n\epsilon^2 &= \log(\delta/2)\\
    \implies n &= \frac{\log(2/\delta)}{2\epsilon^2}\\
    \implies |\widehat{q} - q| &= \sqrt{\frac{\log(2/\delta)}{2n}}.
\end{align*}
Now, we show that with high probability, the estimate for $1 - r^{s\rightarrow t} = \frac{q_t}{q_s}$ is close to the true value. This part of the derivation is similar to that used in \cite{garg2021mixture}. Using triangle inequality,
\begin{align*}
    \left| \frac{\widehat{q}_t}{\widehat{q}_s} - \frac{q_t}{q_s} \right|
    &= \left| \frac{q_s \widehat{q}_t - \widehat{q}_s q_t}{\widehat{q}_s q_s} \right|\\
    &= \frac{1}{\widehat{q}_s q_s}\left| q_s \widehat{q}_t - q_s q_t + q_s q_t - \widehat{q}_s q_t \right|\\
    &\leq \frac{1}{\widehat{q}_s q_s}\left| q_s \widehat{q}_t - q_s q_t\right| + \frac{1}{\widehat{q}_s q_s} \left|q_s q_t - \widehat{q}_s q_t \right|\\
    &\leq \frac{1}{\widehat{q}_s}\left| \widehat{q}_t - q_t\right| + \frac{q_t}{\widehat{q}_s q_s} \left|q_s - \widehat{q}_s \right|.
\end{align*}
On the right hand side, we use the union bound and plug in $\delta / 2$ for $\delta$ in Hoeffding's bound. Plugging in, we then have that with probability at least $1-\delta$,
\begin{align*}
    \left| \frac{\widehat{q}_t}{\widehat{q}_s} - \frac{q_t}{q_s} \right| &\leq \frac{1}{\widehat{q}_s}\left(\sqrt{\frac{\log(4/\delta)}{2n_t}} + \frac{q_t}{q_s} \sqrt{\frac{\log(4/\delta)}{2n_s}}\right)\\
    \implies \left| \widehat{r}^{s\rightarrow t} - r^{s\rightarrow t} \right|
    &\leq \frac{1}{\widehat{P}^s(\widetilde{x} = 1)}\left(\sqrt{\frac{\log(4/\delta)}{2n_t}} + (1-r^{s\rightarrow t}) \sqrt{\frac{\log(4/\delta)}{2n_s}}\right).
\end{align*}

%% file: appendix/A70_nonparametric_justification.tex
\paragraph{Simple Justification} Since \eqref{eqn:s_to_t_elementwise} matches the form of \eqref{eqn:elementwise_corruption} except with $m=r^{s \rightarrow t}$, applying missingness with rate $r^{s\rightarrow t}$ to the source distribution will yield samples independent and identically distributed to the target distribution. That is, plugging in $\widetilde{p}^s$ for $p$ and $r^{s\rightarrow t}$ for $m$, we have:
\begin{align*}
    \widetilde{p}_{x,y} 
    &= \sum_{z: z \leadsto x} p_{z, y} \cdot \prod_{j=1}^d (1-m_{j})^{[x_j]_{\neq 0}} m_{j}^{[z_j]_{\neq 0} - [x_j]_{\neq 0}}\\
    &= \sum_{z: z \leadsto x} \widetilde{p}^s_{z, y} \cdot \prod_{j=1}^d (1-r^{s\rightarrow t}_{j})^{[x_j]_{\neq 0}} (r^{s\rightarrow t}_{j})^{[z_j]_{\neq 0} - [x_j]_{\neq 0}}\\
    &= \widetilde{p}^t_{x,y}
\end{align*}
where the first line is $\eqref{eqn:elementwise_corruption}$ and the third line follows from $\eqref{eqn:s_to_t_elementwise}$.

\paragraph{Alternative Justification}
Suppose that $m^t \succeq m^s$, where $\succeq$ denotes whether all elements of $m^t$ are greater than or equal to all corresponding elements of $m^s$, that is, $m_j^t \geq m_j^s$ for $j = 1, 2, ..., d$. Below, we show that the data generating process for the target data is equivalent to applying a missingness filter with relative missingness rate $r^{s\rightarrow t}$ applied to the source data. To draw a point from the source, target, and transformed distribution, respectively, one first draws a clean data point $(x, y) \sim P(X, Y)$, where $x \in \mathbb{R}^d, y \in \mathbb{R}$, and then applies the respective missingness filter to the clean covariates:
\begin{align*}
    \widetilde{x}^s &= \nu_{s}(x) = x \odot \xi^{s}\\
    \widetilde{x}^t &= \nu_{t}(x) = x \odot \xi^{t}\\
    \widetilde{x}^{s \rightarrow t} &= \nu_{s \rightarrow t}(\nu_{s}(x)) = x \odot \xi^{s} \odot \xi^{s \rightarrow t}
\end{align*}
where $\xi^t \sim \text{Bernoulli}(1 - m^t)$, $\xi^s \sim \text{Bernoulli}(1 - m^s)$, and $\xi^{s \rightarrow t} \sim \text{Bernoulli}(1 - r^{s \rightarrow t})$. Combining Bernoullis, we have:
\begin{align*}
    \xi^{s} \odot \xi^{s\rightarrow t} 
&=\left\{\begin{matrix}
1 & \text{w.p. } \left(1 - \frac{m^t - m^s}{1-m^{s}}\right) \cdot (1 - m^s) \\
0 &  otherwise \\
\end{matrix}\right.\\
&=\left\{\begin{matrix}
1 & \text{w.p. } (1 - m^{t}) \\
0 &  otherwise \\
\end{matrix}\right. 
=\xi^t
\end{align*}

Thus, for true relative missing rates $r^{s \rightarrow t}$, we have $\nu_{t}(x) = \nu_{s \rightarrow t}(\nu_{s}(x))$. Since the data generating process after applying $\nu_{s\rightarrow t}$ to source data 
is now identical to the data generating process of the target dataset, 
we have
$\{(\nu_{s\rightarrow t}(\widetilde{X}^{s,i}), Y^{s,i})\}_{i=1}^{n_s}$ 
drawn independent and identically distributed to $P^t(\widetilde{X}, Y)$.

%% file: appendix/A80_linear_predictors.tex
\subsection{Optimal linear target predictor, derived from target covariances}\label{app:optimal_linear_from_target}
For each dimension $j$, the covariance between corrupted data $\Xt_j$ with missingness rate $m$ and its labels $Y$ is $\smallCov{\Xt_j, Y} 
= \smallCov{X_j \cdot \xi_j, Y} 
= (1 - m_j) \Cov{X_j, Y}$. Thus,
\begin{align*}
    \smallCov{X, Y} &= \frac{1}{1-m} \odot \smallCov{\Xt, Y}\\
    \smallE{X^\top Y} &= \Cov{X, Y} + \smallE{X}^\top\smallE{Y}\\
    &= \frac{1}{1-m} \odot \Cov{\Xt, Y} + \frac{1}{1-m} \odot \smallE{\Xt}^\top\smallE{Y}\\
    &= \frac{1}{1-m} \odot \smallE{\Xt^\top Y}.
\end{align*}
Plugging into the ordinary least squares regression solution,
\begin{align*}
    \beta^t_* &= \smallE{\Xt^{t^\top} \Xt^t}^{-1}\smallE{\widetilde{X}^{t\top} Y^t}\\
    &= \smallE{\Xt^{t\top} \Xt^t}^{-1}\left((1 - m_t) \odot \smallE{X^\top Y}\right)\\
    &= \smallE{\Xt^{t\top} \Xt^t}^{-1}\left(\frac{1 - m_t}{1-m_s} \odot \smallE{\widetilde{X}^{s\top} Y^s}\right)\\
    &= \smallE{\Xt^{t\top} \Xt^t}^{-1}\left(r^{s\rightarrow t} \odot \smallE{\widetilde{X}^{s\top} Y^s}\right).
\end{align*}

The remainder of this section derives the optimal linear target predictor, where the corrupted target covariance is derived from the corrupted source covariance.

\subsection{Means, Variances, and Covariances}
This section begins by deriving the relationships between the means, covariances, and variances of the corrupted and clean data. Then, it derives the relationships between corrupted and clean $\smallE{X^\top  X}$. 
Finally, the derived first and second moments are summarized in Table \ref{tab:moments}.

Recall that for any covariate $x_j$, we have:
\begin{align*}
    \widetilde{x}_j &= \left\{\begin{matrix}
0 & \text{w.p. } m_j\\ 
x_j & \text{w.p. } 1- m_j
\end{matrix}\right. \\
&= b_j x_j
\end{align*}
where $b_j \sim \text{Bernoulli}(1 - m_j)$. The \textbf{mean} of the corrupted data is given by:
\begin{align*}
    \smallE{\Xt} = (1-m) \odot \E{X}
\end{align*}

To derive the covariance matrix of the corrupted data, consider the covariance between two arbitrary distinct covariate dimensions $\widetilde{x}_1$ and $\widetilde{x}_2$. Let $A = b_1$, $B = x_1$, $C = b_2$, and $D = x_2$. Note that $A$ and $C$ are independent of all other variables. Thus,

\begin{align*}
    \text{Cov}(\widetilde{x}_1, \widetilde{x}_2) &= \text{Cov}(AB, CD)\\
    &= \mathbb{E}[ABCD] - \mathbb{E}[AB]\mathbb{E}[CD]\\
    &= \mathbb{E}[ABCD] - \mathbb{E}[A]\mathbb{E}[B]\mathbb{E}[C]\mathbb{E}[D]\\
    &= \mathbb{E}[A]\mathbb{E}[C] (\mathbb{E}[BD] - \mathbb{E}[B]\mathbb{E}[D])\\
    &= \mathbb{E}[A]\mathbb{E}[C] \text{Cov}(B, D)\\
    &= (1 - m_1)(1 - m_2)\text{Cov}\left (x_1, x_2 \right )\\
    \implies \text{Cov}(x_1, x_2) 
    &= \frac{1}{(1 - m_1)(1 - m_2)} \text{Cov}(\widetilde{x}_1, \widetilde{x}_2)
\end{align*}
And similarly, 
\begin{align*}
    \text{Cov}(\widetilde{x}_1, y) &= (1-m_1) \Cov{x_1, y}\\
    \implies \Cov{x_1, y} &= \frac{1}{1-m_1}\text{Cov}(\widetilde{x}_1, y)
\end{align*}

The variance (entries along the diagonal of the covariance matrix) is given by:
\begin{align*}
    \text{Var}(\widetilde{x}_1) &= \text{Var}\left (b_1 x_1 \right )\\
    &=\text{Var}(AB)\\
    &= (\sigma_A^2 + \mu_A^2)(\sigma_B^2 + \mu_B^2) - \mu_A^2 \mu_B^2\\
    &= (m_1(1-m_1) + (1 - m_1)^2)\left(\text{Var}(x_1) + \mathbb{E}[x_1]^2\right) - (1 - m_1)^2 \mathbb{E}[x_1]^2\\
    &= (1 - m_1)\left(\text{Var}(x_1) + \mathbb{E}[x_1]^2\right) - (1 - m_1)^2 \mathbb{E}[x_1]^2\\
    &= (1 - m_1) \left (\text{Var}(x_1) + \mathbb{E}[x_1]^2 - (1 - m_1) \mathbb{E}[x_1]^2\right )\\
    &= (1 - m_1) \left (\text{Var}(x_1) + \mathbb{E}[x_1]^2 - \mathbb{E}[x_1]^2 + m_1\mathbb{E}[x_1]^2\right )\\
    &= (1 - m_1) \left (\text{Var}(x_1) + m_1\mathbb{E}[x_1]^2\right )\\
    &= (1 - m_1)\text{Var}(x_1) + m_1(1- m_1) \mathbb{E}[x_1]^2\\
    \text{Var}(x_1) &= \frac{\text{Var}(\widetilde{x}_1)}{1-m_1} - m_1 \mathbb{E}[x_1]^2\\
    &= \frac{\text{Var}(\widetilde{x}_1)}{1-m_1} - \frac{m_1}{(1-m_1)^2} \mathbb{E}[\widetilde{x}_1]^2\\
\end{align*}

Putting this together, the variance-covariance matrix is given by (elementwise division below):
{\small
\begin{align*}
    \text{Cov}(\widetilde{X},\widetilde{X}) 
    &= (1 - m)(1 - m)^\top  \odot \text{Cov}(X,X) \\
    &\hspace{3em}+ \text{diag}(((1 - m) - (1 - m)^2)\text{Var}(X) + m(1-m)\mathbb{E}[x_1]^2)\\
    &= (1 - m)(1 - m)^\top  \odot \text{Cov}(X,X) 
    + \text{diag}(m (1-m) (\text{Var}(X) + \mathbb{E}[X]^2))\\
    &= (1 - m)(1 - m)^\top  \odot \text{Cov}(X,X) \\
    &\hspace{3em}+ \text{diag}(m (1-m)^\top )\text{diag}(\text{Cov}(X,X) + \mathbb{E}[X]^\top \mathbb{E}[X])\\
    &= (1 - m)(1 - m)^\top  \odot \text{Cov}(X,X) + \text{diag}(m (1-m)^\top )\text{diag}(\mathbb{E}[X^\top X])\\
    \implies \text{Cov}(X,X) 
    &= \left(\frac{1}{1 - m}\right)\left(\frac{1}{1 - m}\right)^\top  \odot \text{Cov}(\Xt,\Xt) \\
    &\hspace{3em}+ \diag{-\frac{\smallVar{\Xt}}{(1-m)^2} + \frac{\smallVar{\Xt}}{1-m} - \frac{m\E{\Xt}^2}{(1-m)^2}}\\
    &= \left(\frac{1}{1 - m}\right)\left(\frac{1}{1 - m}\right)^\top  \odot \text{Cov}(\Xt,\Xt) - \diag{\frac{m}{(1-m)^2}(\smallVar{\Xt} + \smallE{\Xt}^2)}
\end{align*}
}%
Thus far, we have been working with the covariance matrix. How do the expressions for covariance relate to $\widetilde{X}^\top \widetilde{X}$ and $\widetilde{X}^\top  Y$? We have:
\begin{align*}
    \text{Cov}(\widetilde{X}, \widetilde{X}) &= (1-m)(1-m)^\top  \odot \text{Cov}(X,X) + \text{diag}\left( m(1-m)^\top  \right) \text{diag}(\mathbb{E}[X^\top X])\\
    \smallE{\Xt^\top \Xt}&= \Cov{\Xt, \Xt} + \smallE{\Xt}^\top \smallE{\Xt}\\
    &= (1-m)(1-m)^\top  \odot (\smallCov{X, X} + \smallE{X}^\top \smallE{X}) \diag{m(1-m)^\top }\diag{\E{X^\top X}}\\
    &= (1-m)(1-m)^\top  \odot \E{X^\top X} + \diag{m(1-m^\top )}\diag{\E{X^\top X}}
\end{align*}
Additionally,
{\small
\begin{align*}
    \Cov{X,X} &= \left(\frac{1}{1 - m}\right)\left(\frac{1}{1 - m}\right)^\top  \odot \text{Cov}(\Xt,\Xt) + 
    \diag{-\frac{m}{(1-m)^2}}
    \diag{\smallVar{\Xt} + \smallE{\Xt}^2}\\
    \E{X^\top X} &= \Cov{X,X} + \E{X}^\top \E{X}\\
    &= \left(\frac{1}{1 - m}\right)\left(\frac{1}{1 - m}\right)^\top  \odot \left(\text{Cov}(\Xt,\Xt) + \smallE{\Xt}^\top \smallE{\Xt}\right) \\
    &\hspace{3em}+
    \diag{-\frac{m}{(1-m)^2}}
    \diag{\smallVar{\Xt} + \smallE{\Xt}^2}\\
    &= \left(\frac{1}{1 - m}\right)\left(\frac{1}{1 - m}\right)^\top  \odot \smallE{\Xt^\top \Xt} - \diag{\frac{m}{(1-m)^2}}
    \diag{\smallE{\Xt^\top \Xt}}
\end{align*}
}%

\begin{table}[htbp]
    \centering
    \caption{Summary of 1st and 2nd moments of corrupted data and clean data}
    \label{tab:moments}
    \renewcommand{\arraystretch}{1.5}
    \setlength{\tabcolsep}{12pt}
    \hspace*{-3cm}\begin{tabular}{cc}
        \toprule
        Quantity of Interest & Expression \\
        \midrule
        $\E{X}$ &  $\frac{1}{1-m} \odot \E{\Xt}$\\
        $\E{\Xt}$ & $(1-m) \odot \E{X}$\\
        $\E{X^\top X}$ & $\left(\frac{1}{1 - m}\right)\left(\frac{1}{1 - m}\right)^\top  \odot \E{\Xt^\top \Xt} - \diag{\frac{m}{(1-m)^2}}
    \diag{\smallE{\Xt^\top \Xt}}$ \\
        $\E{\Xt^\top \Xt}$ & $(1-m)(1-m)^\top  \odot \E{X^\top X} + \diag{m(1-m)^\top }\diag{\E{X^\top X}}$\\
        \bottomrule
    \end{tabular}\hspace*{-3cm}
\end{table}

\subsection{Closed Form Solution}
Using results from previous sections, we can now derive a closed form solution for the optimal linear classifier for a target domain with missing rates $m_t$, given labeled data from a source domain with missing rates $m_s$. We break down this problem by going from corrupted data with some missingness rate to clean data with 0 missingness, and then from clean data with 0 missingness to corrupted data with another level of missingness.

Suppose we are going from corrupted data $\Xt$ with missing rate $m$ to clean data $X$ with 0 missingness:
{\small
\begin{align*}
    \Cov{X, y} &= \frac{1}{1-m} \odot \Cov{\Xt, y}\\
    \E{X^\top y} &= \Cov{X, y} + \E{X}^\top \E{y}\\
    &= \frac{1}{1-m} \odot \Cov{\Xt, y} + \frac{1}{1-m} \odot \E{\Xt}^\top \E{y}\\
    &= \frac{1}{1-m} \odot \E{\Xt^\top  y}\\
    \E{X^\top X} &= \left(\frac{1}{1-m}\right) \left(\frac{1}{1-m}\right)^\top  \odot  \E{\Xt^\top  \Xt} - \diag{\frac{m}{(1-m)^2} \odot \E{\Xt^\top  \Xt}}\\
    \implies \beta  &= \left\{\left(\frac{1}{1-m}\right)\left(\frac{1}{1-m}\right)^\top  \odot \E{\Xt^\top  \Xt} - \diag{\frac{m}{(1-m)^2} \odot \E{\Xt^\top  \Xt}}\right\}^{-1} \frac{1}{1-m} \odot \E{\Xt^\top  y}
\end{align*}
}%
Going from clean to corrupted data, we have:
{\small
\begin{align*}
    \smallE{\Xt^\top  y} 
    &= \smallCov{\Xt, y} + \smallE{\Xt}^\top  \E{y}\\
    &= (1-m) \odot \Cov{X, y} + (1-m) \odot \E{\Xt}^\top  \E{y}\\
    \smallE{\Xt^\top \Xt} &= (1-m)(1-m)^\top  \odot \E{X^\top  X} + \diag{m(1-m^\top )}\diag{\E{X^\top X}}\\
    \implies \widetilde{\beta} &=\left[(1-m)(1-m)^\top  \odot \E{X^\top  X} + \diag{m(1-m^\top )} \diag{\E{X^\top X}}\right]^{-1} (1-m) \odot \E{X^\top  y}
\end{align*}
}%

Now, we put all of these equations together, going from source corrupted data (S), to clean data (C), to target corrupted data (T).

\noindent (S) $\rightarrow$ (C):
{\small
\begin{align*}
    \smallE{X^\top  X} &= \left(\frac{1}{1-m_s}\right)\left(\frac{1}{1-m_s}\right)^\top  \odot \smallE{\Xt^{s\top}  \Xt^s} - \diag{\frac{m_s}{(1-m_s)^2}\odot (\mathbb{E}[\Xt^{s\top} \Xt^s])}\\
    \smallE{X^\top  y} &= \frac{1}{1-m_s}\odot \Cov{\Xt^s, y} + \frac{1}{1-m_s}\odot\smallE{\Xt^s}^\top \E{y}
\end{align*}
}%

\noindent (C) $\rightarrow$ (T):
{\small
\begin{align*}
    \E{\Xt^{t\top}  \Xt^{t}} &= (1-m_t)(1-m_t)^\top  \odot \E{X^\top X} + \diag{m_t(1-m_t^\top )}\diag{\E{X^\top X}}\\
    &= (1-m_t)(1-m_t)^\top  \odot \left[\left(\frac{1}{1-m_s}\right)\left(\frac{1}{1-m_s}\right)^\top  \odot \E{\Xt^{s\top} \Xt^s} - \diag{\frac{m_s}{(1-m_s)^2}\E{\Xt^{s\top} \Xt^s}}\right]\\
    &\text{\hspace{2em}} + \diag{\frac{m_t}{1-m_t}} \odot \diag{\E{\Xt^{s\top} \Xt^s} - \diag{m_s}\E{\Xt^{s\top}  \Xt^s}}\\
    &= (1-m_t)(1-m_t)^\top  \odot \left(\frac{1}{1-m_s}\right)\left(\frac{1}{1-m_s}\right)^\top  \odot \E{\Xt^{s\top} \Xt^s}\\
    &\text{\hspace{2em}}- (1-m_t)(1-m_t)^\top  \odot \diag{\frac{m_s}{(1-m_s)^2} \odot \E{\Xt^{s\top} \Xt^s}}\\
    &\text{\hspace{2em}} + \diag{\frac{m_t(1-m_t)}{1-m_s} \odot \E{\Xt^{s\top}  \Xt^s}}\\
\end{align*}
}%
For $i \neq j$, the off-diagonal entries of the above expression are given by:
{\small
\begin{align*}
    \E{\Xt^{t\top}  \Xt^{t}}_{ij} 
    = \left (\frac{1-m_{ti}}{1-m_{si}}\right )\left (\frac{1-m_{tj}}{1-m_{sj}}\right ) \E{\Xt^{s\top}  \Xt^s}_{ij}
    = (1-r^{s\rightarrow t}_i)(1-r^{s\rightarrow t}_j)\E{\Xt^{s\top}  \Xt^s}_{ij}
\end{align*}
}%

The diagonal entries of the above expression are given by:
{\small
\begin{align*}
    \E{\Xt^{t\top}  \Xt^{t}}_{ii} 
    &= \E{\Xt^{s\top}  \Xt^s}_{ii} \left (
    \left (\frac{1-m_{ti}}{1-m_{si}}\right )^2  
    - \frac{m_{si}(1-m_{ti})^2}{(1-m_{si})^2}
    + \frac{m_{ti}(1-m_{ti})}{1-m_{si}}
    \right )\\
    &= \E{\Xt^{s\top}  \Xt^s}_{ii} \left (
    (1-r^{s\rightarrow t}_i)^2  
    -  m_{si} (1-r^{s\rightarrow t}_i)^2
    + m_{ti}(1-r^{s\rightarrow t}_i)
    \right )\\
    &= \E{\Xt^{s\top}  \Xt^s}_{ii} (1-r^{s\rightarrow t}_i) \left (
    (1-r^{s\rightarrow t}_i)
    -  m_{si} (1-r^{s\rightarrow t}_i)
    + m_{ti}
    \right )\\
    &= \E{\Xt^{s\top}  \Xt^s}_{ii} (1-r^{s\rightarrow t}_i) \left (
    \frac{1 - m_{ti}}{1-m_{si}}
    -  \frac{m_{si} - m_{si}m_{ti}}{1-m_{si}}
    + \frac{m_{ti}-m_{si}m_{ti}}{1-m_{si}}
    \right )\\
    &= \E{\Xt^{s\top}  \Xt^s}_{ii} (1-r^{s\rightarrow t}_i) \left (
    \frac{1 - m_{si}}{1-m_{si}}
    \right )\\
    &= \E{\Xt^{s\top}  \Xt^s}_{ii} (1-r^{s\rightarrow t}_i)
\end{align*}
}%

Additionally,
{\small
\begin{align*}
    \E{\Xt^{t\top}  y} 
    &= (1-m_t) \odot \E{X^\top  y} \\
    &= (1-m_t) \odot \left( \frac{1}{1-m_s} \odot \Cov{\Xt^s, y} + \frac{1}{1-m_s} \odot \E{\Xt^s}^\top \E{y} \right) \\
    &= \frac{1-m_t}{1-m_s} \odot \E{\Xt^{s\top}  y}
\end{align*}
}%

%% file: appendix/A90_experiment_details.tex
Experiments were run on a machine with 28 CPU cores. The linear regression models were implemented from scratch and validated against that of sklearn. The MLPRegressor class from the scikit-learn Python package was used with default hyperparameters, and the XGBoost class from the xgboost Python package was used with default hyperparameters. All experiments (except imputation) are feasible to run within a few hours.

Semi-synthetic experiments on linear models included 10 samples of $\beta$, and 50 samples of missingness rates under each regime ($m^s \preceq m^t$ and $m^s \text{ ? } m^t$). Semi-synthetic experiments on nonlinear models (XGB, NN) included 5 samples of $\beta$ and 20 samples of missingness rates under each regime. Across these runs, 95\% confidence intervals were computed.

In the imputation experiments, a MissForest imputer from the missingpy Python package was trained on the combination of the source training set and target training set (just on the covariates, without labels). This imputer was then applied to both the source and target test sets. Finally, we train a source classifier on the imputed source labeled data and evaluate its performance on the target unlabeled data. We note that in our experience with the imputation experiments, imputation was somewhat slow (2-3 minutes for each imputation), and so all of our imputed results are reported on 5 samples of $\beta$ and 20 samples of missingness rates under each regime, across all semi-synthetic datasets.

\subsection{Synthetic Data Experiments}
\begin{table}[h]
    \centering
    \caption{MSE/Var(Y) on Redundant Features and Confounded Features settings, with 95\% confidence intervals computed over varying $\epsilon$ between 0.05 to 0.95.}
    \label{app:synthetic_experiments}
    \input{tables/ci_synthetic}
\end{table}

\subsection{Semi-Synthetic Data Experiments}
\label{app:preprocess_semisynthetic}
The UCI datasets \cite{Dua:2019_UCI} used in this work are:
\begin{itemize}
    \item Adult Data Set: The classification task is whether an individual's income exceeds \$50K a year based on census data. The dataset contains categorical variables (occupation, education, marital status, etc.), as well as continuous variables (age, hours per week, etc.)
    \item Bank Marketing Data Set: The classification task is whether a client will subscribe a term deposit. This dataset contains categorical features such as type of job, marital status, education, whether they have a housing loan, etc., as well as continuous variables such as age, number of contacts performed, etc.
    \item Thyroid Disease Data Set: The classification task is of increased vs. decreased binding protein. This dataset contains binary variables such as whether the patient is pregnant, is male, on thyroxine, has a tumor, etc., as well as continuous variables such as age, TSH, T3, TT4, etc.
\end{itemize}
For semi-synthetic experiments, we pre-process the UCI data by creating dummy variables from categorical variables, dropping redundant columns, normalizing numerical variables, dropping binary variables with low frequency ($<5\%$, since we apply additional synthetic missingness in our experiments), and dropping columns with low variance ($<5\%$). We additionally generate synthetic labels by sampling coefficients $\beta_j \sim \text{Uniform}(0, 10), \forall j \in \{0, 1, 2, ..., d\}$ and computing new synthetic labels $y_{new} = X\beta$. Table \ref{app:adult_conf} contains the MSE/Var(Y) and 95\% confidence intervals (from sampling several $\beta$ and $m^s, m^t$) of the adult dataset, Table \ref{app:bank_conf} contains the MSE/Var(Y) and 95\% confidence intervals of the bank dataset, and Table \ref{app:thyroid_conf} contains the MSE/Var(Y) and 95\% confidence intervals of the thyroid dataset.

\begin{table}[H]
    \caption{MSE/Var(Y) on UCI Adult Semi-synthetic Setting, with 95\% confidence intervals computed over multiple samples of $\beta$ and $m^s, m^t$ (described in Section \ref{sec:experiments}).} \label{app:adult_conf}
    \centering
    \input{tables/ci_adult}
\end{table}

\begin{table}[H]
    \caption{MSE/Var(Y) on UCI Bank Semi-synthetic Setting, with 95\% confidence intervals computed over multiple samples of $\beta$ and $m^s, m^t$ (described in Section \ref{sec:experiments}).}
    \label{app:bank_conf}
    \centering
\input{tables/ci_bank}
\end{table}

\begin{table}[H]
    \caption{MSE/Var(Y) on UCI Thyroid Semi-synthetic Setting, with 95\% confidence intervals computed over multiple samples of $\beta$ and $m^s, m^t$ (described in Section \ref{sec:experiments}).}
    \label{app:thyroid_conf}
    \centering
\input{tables/ci_thyroid}
\end{table}

\subsection{Real Data Experiments}
\label{app:preprocess_eicu}
The data for these experiments were derived from eICU-CRD \citep{pollard2018eicu}, a multi-hospital critical care database which uses the PhysioNet Credentialed Health Data License Version 1.5.0. We extract data for predicting 48-hour mortality through the FIDDLE \citep{tang2020fiddle} preprocessing pipeline with default parameters. FIDDLE extracts both time-varying and fixed features. We collapse the time-varying features by taking the maximum value (note that most features are binary, and none take values less than 0). We extract data from two of the hospitals with the most data, the first of which contains 3,006 data points, and the second of which contains 2,663 data points. The rate of 48-hour mortality in the first hospital is 0.097, and the rate of 48-hour mortality in the second hospital is 0.100. Additionally, we threshold for features that are present that have a prevalence of at least 5\% in either of the hospitals and at least 1\% in both of the hospitals. Code is provided at \href{https://github.com/acmi-lab/Missingness-Shift}{https://github.com/acmi-lab/Missingness-Shift}. We used target unlabeled data ($\alpha_t = 1, \alpha_s = 0$) to estimate $\smallE{\widetilde{X}^{t\top} \widetilde{X}^t}$ for the adjusted linear closed form model because we noticed that the estimation error with limited data made the source estimates less reliable. 
Due to limited positive samples, in order to evaluate cross-domain performance, a model was trained on all data from one domain and tested on all data from the other. Oracle performance (training and testing on the same domain) was computed from training on a randomly sampled 80\% of the data and testing on the remaining 20\%.
Table \ref{app:est_r} contains the estimated relative non-missingness of the top five coefficients for the oracle models from each hospital.

\begin{table}[h]
    \centering
    \caption{The estimated proportion of nonzeros in Hospital 1 ($q_1$) and Hospital 2 ($q_2$), estimated relative non-missingness rates $q_2/q_1 = 1 - r^{1\rightarrow 2}$, Hospital 1 Oracle coefficient ($\beta_1$), and Hospital 2 Oracle coefficient ($\beta_2$) for each of the top five features (measure by magnitude of coefficient) from the Oracle linear predictors of Hospital 1 and 2.}
    \label{app:est_r}
\begin{tabular}{lrrrrr}
\toprule
{} & $\beta_1$ &  $\beta_2$ & $q_1$ &  $q_2$ &  $q_2/q_1$ \\
\midrule
noninvasivemean\_max\_(78.0, 86.0]                                                              & -0.279 & -0.364 &  0.754 &  0.938 &  1.244 \\
systemicsystolic\_mean\_(-94.001, 99.667]                                                       &  0.271 & -0.362 &  0.333 &  0.134 &  0.404 \\
unittype...Neuro ICU                                                                      &  0.055 & -0.577 &  0.194 &  0.315 &  1.629 \\
ethnicity...African American                                                              & -0.275 &  0.361 &  0.141 &  0.071 &  0.506 \\
...Intake (ml)...(100.0, 150.0] &  0.070 & -0.732 &  0.318 &  0.045 &  0.142 \\
...Invasive BP Systolic...(-59.001, 101.0]                           & -0.571 &  0.474 &  0.350 &  0.130 &  0.372 \\
cvp\_max\_(8.0, 12.0]                                                                           &  0.536 & -0.476 &  0.262 &  0.125 &  0.477 \\
\bottomrule
\end{tabular}
    \label{tab:eicu_nonmissingness}
\end{table}

%% file: tables/ci_synthetic.tex
\begin{tabular}{lcc}
\toprule
{} &                     $m^s \preceq m^t$ &                     $m^s \text{ ? } m^t$ \\
\midrule
Lin. Reg. (oracle)           &  \textcolor{gray}{0.178 (0.172 -- 0.185)} &  \textcolor{gray}{0.206 (0.199 -- 0.213)} \\
Lin. Reg. (source)           &                    1.259 (1.231 -- 1.286) &                    1.103 (1.076 -- 1.129) \\
Lin. Reg. (imputed)          &                    1.002 (1.002 -- 1.002) &                    0.918 (0.915 -- 0.921) \\
Lin. Reg. (closed-form adj.) &           \textbf{0.186 (0.180 -- 0.193)} &           \textbf{0.209 (0.205 -- 0.213)} \\
Lin. Reg. (non-param. adj.)  &                    0.473 (0.471 -- 0.476) &                    0.492 (0.489 -- 0.495) \\
\midrule
XGBoost (oracle)           &  \textcolor{gray}{0.166 (0.160 -- 0.172)} &  \textcolor{gray}{0.200 (0.193 -- 0.208)} \\
XGBoost (source)           &                    \textbf{0.166 (0.160 -- 0.172)} &                    0.475 (0.458 -- 0.492) \\
XGBoost (imputed)          &                             1.002 (1.002 -- 1.002) &                    1.157 (1.102 -- 1.211) \\
XGBoost (non-param. adj.)  &                             0.425 (0.422 -- 0.428) &           \textbf{0.473 (0.468 -- 0.478)} \\
\midrule
MLP (oracle)           &  \textcolor{gray}{0.166 (0.160 -- 0.172)} &  \textcolor{gray}{0.201 (0.195 -- 0.208)} \\
MLP (source)           &           \textbf{0.184 (0.165 -- 0.202)} &           \textbf{0.321 (0.300 -- 0.342)} \\
MLP (imputed)          &                    1.003 (1.002 -- 1.003) &                    0.924 (0.918 -- 0.930) \\
MLP (non-param. adj.)  &                    0.436 (0.428 -- 0.444) &                    0.470 (0.465 -- 0.474) \\
\bottomrule
\end{tabular}

%% file: tables/ci_adult.tex
\begin{tabular}{lll}
\toprule
{} &                     $m^s \preceq m^t$ &                     $m^s \text{ ? } m^t$ \\
\midrule
Lin. Reg. (oracle)           &  \textcolor{gray}{0.420 (0.415 -- 0.424)} &           \textcolor{gray}{0.362 (0.356 -- 0.367)} \\
Lin. Reg. (source)           &           0.437 (0.433 -- 0.442) &           0.380 (0.373 -- 0.386) \\
Lin. Reg. (imputed)          &           0.490 (0.471 -- 0.509) &           0.483 (0.475 -- 0.491) \\
Lin. Reg. (closed-form adj.) &           0.422 (0.417 -- 0.426) &  \textbf{0.363 (0.358 -- 0.368)} \\
Lin. Reg. (non-param. adj.)  &  \textbf{0.420 (0.415 -- 0.424)} &           0.373 (0.367 -- 0.379) \\
\midrule
XGBoost (oracle)           &           \textcolor{gray}{0.398 (0.386 -- 0.409)} &           \textcolor{gray}{0.354 (0.344 -- 0.363)} \\
XGBoost (source)           &           0.399 (0.387 -- 0.410) &           0.379 (0.369 -- 0.388) \\
XGBoost (imputed)          &           0.512 (0.491 -- 0.534) &           0.521 (0.508 -- 0.535) \\
XGBoost (non-param. adj.)  &           0.399 (0.387 -- 0.410) &           0.392 (0.382 -- 0.402) \\
\midrule
MLP (oracle)               &  \textcolor{gray}{0.389 (0.378 -- 0.401)} &           \textcolor{gray}{0.343 (0.334 -- 0.352)} \\
MLP (source)               &           0.399 (0.387 -- 0.410) &           0.357 (0.348 -- 0.367) \\
MLP (imputed)              &           0.480 (0.461 -- 0.499) &           0.468 (0.456 -- 0.481) \\
MLP (non-param. adj.)      &  \textbf{0.389 (0.378 -- 0.400)} &  \textbf{0.355 (0.346 -- 0.364)} \\
\bottomrule
\end{tabular}

%% file: tables/ci_bank.tex
\begin{tabular}{lll}
\toprule
{} &                     $m^s \preceq m^t$ &                     $m^s \text{ ? } m^t$ \\
\midrule
Lin. Reg. (oracle)           &  \textcolor{gray}{0.338 (0.336 -- 0.340)} &           \textcolor{gray}{0.433 (0.426 -- 0.440)} \\
Lin. Reg. (source)           &           0.371 (0.369 -- 0.373) &           0.480 (0.472 -- 0.487) \\
Lin. Reg. (imputed)          &           0.501 (0.491 -- 0.511) &           0.592 (0.583 -- 0.602) \\
Lin. Reg. (closed-form adj.) &           0.339 (0.337 -- 0.340) &  \textbf{0.442 (0.436 -- 0.449)} \\
Lin. Reg. (non-param. adj.)  &  \textbf{0.338 (0.336 -- 0.340)} &           0.459 (0.453 -- 0.466) \\
\midrule
XGBoost (oracle)           &  \textcolor{gray}{0.287 (0.279 -- 0.295)} &           \textcolor{gray}{0.453 (0.438 -- 0.468)} \\
XGBoost (source)           &           0.305 (0.297 -- 0.313) &           \textbf{0.500 (0.484 -- 0.516)} \\
XGBoost (imputed)          &           0.492 (0.482 -- 0.503) &           0.708 (0.684 -- 0.732) \\
XGBoost (non-param. adj.)  &  \textbf{0.287 (0.279 -- 0.295)} &           0.503 (0.486 -- 0.519) \\
\midrule
MLP (oracle)               &           \textcolor{gray}{0.295 (0.287 -- 0.303)} &           \textcolor{gray}{0.458 (0.442 -- 0.473)} \\
MLP (source)               &           0.322 (0.314 -- 0.330) &           0.499 (0.483 -- 0.516) \\
MLP (imputed)              &           0.484 (0.474 -- 0.494) &           0.668 (0.645 -- 0.690) \\
MLP (non-param. adj.)      &           \textbf{0.294 (0.286 -- 0.302)} &  \textbf{0.487 (0.471 -- 0.503)} \\
\bottomrule
\end{tabular}

%% file: tables/ci_thyroid.tex
\begin{tabular}{lll}
\toprule
{} &                     $m^s \preceq m^t$ &                     $m^s \text{ ? } m^t$ \\
\midrule
Lin. Reg. (oracle)           &           \textcolor{gray}{0.298 (0.292 -- 0.303)} &           \textcolor{gray}{0.251 (0.246 -- 0.256)} \\
Lin. Reg. (source)           &           0.350 (0.342 -- 0.357) &           0.320 (0.314 -- 0.326) \\
Lin. Reg. (imputed)          &           0.306 (0.298 -- 0.313) &           0.358 (0.351 -- 0.365) \\
Lin. Reg. (closed-form adj.) &           0.316 (0.310 -- 0.322) &  \textbf{0.291 (0.286 -- 0.295)} \\
Lin. Reg. (non-param. adj.)  &  \textbf{0.293 (0.288 -- 0.298)} &  \textbf{0.291 (0.286 -- 0.296)} \\
\midrule
XGBoost (oracle)           &           \textcolor{gray}{0.316 (0.304 -- 0.328)} &           \textcolor{gray}{0.274 (0.265 -- 0.282)} \\
XGBoost (source)           &           \textbf{0.310 (0.298 -- 0.322)} &           \textbf{0.352 (0.341 -- 0.362)} \\
XGBoost (imputed)          &           0.355 (0.346 -- 0.364) &           0.441 (0.430 -- 0.452) \\
XGBoost (non-param. adj.)  &           \textbf{0.310 (0.298 -- 0.321)} &           0.381 (0.370 -- 0.392) \\
\midrule
MLP (oracle)               &           \textcolor{gray}{0.279 (0.269 -- 0.288)} &           \textcolor{gray}{0.230 (0.223 -- 0.236)} \\
MLP (source)               &           0.320 (0.308 -- 0.331) &           0.303 (0.294 -- 0.311) \\
MLP (imputed)              &           0.304 (0.296 -- 0.311) &           0.345 (0.336 -- 0.355) \\
MLP (non-param. adj.)      &  \textbf{0.278 (0.268 -- 0.288)} &  \textbf{0.272 (0.265 -- 0.279)} \\
\bottomrule
\end{tabular}